%% file: main.tex
\documentclass[10pt,twocolumn,letterpaper]{article}

\usepackage{cvpr}              

\input{preamble}
\definecolor{cvprblue}{rgb}{0.21,0.49,0.74}
\usepackage[pagebackref,breaklinks,colorlinks,allcolors=cvprblue]{hyperref}

\def\paperID{43914} 
\def\confName{CVPR}
\def\confYear{2026}

\title{Free-Lunch Long Video Generation via Layer-Adaptive O.O.D Correction}

\author{Jiahao Tian \quad \quad \quad  Chenxi Song\textsuperscript{*} \quad \quad  \quad Wei Cheng \quad \quad  \quad Chi Zhang\textsuperscript{$\dagger$}  \\[0.4em]
Westlake University \\
}

\begin{document}
\maketitle
{
  \renewcommand{\thefootnote}%
    {\fnsymbol{footnote}}
 \footnotetext{\hspace*{-1em}* denotes project lead. $\dagger$ denotes corresponding author.
  \\  \hspace*{1em} Emails: \texttt{\{tianjiahao, chizhang\}@westlake.edu.cn}
  }
}
\addtocontents{toc}{\protect\setcounter{tocdepth}{0}}
\input{sec/0_abstract}    
\input{sec/1_intro}

\input{sec/2_related_work}

\input{sec/3_method}
\input{sec/4_experiments}

\input{sec/5_conclusion}
\section*{Acknowledgement}
This work was supported by the National Natural Science Foundation of China (No. 6250070674) and the Zhejiang Leading Innovative and Entrepreneur Team Introduction Program (2024R01007).
{
    \small
    \bibliographystyle{ieeenat_fullname}
    \bibliography{main}
}

\input{sec/X_suppl}


\end{document}

%% file: preamble.tex
\definecolor{mitblue}{rgb}{0.88,0.95,0.96}

\usepackage[utf8]{inputenc}
\usepackage[T1]{fontenc}

\usepackage{color,xcolor}
\usepackage{epsfig}
\usepackage{graphicx}
\usepackage{bm}

\usepackage{adjustbox}
\usepackage{array}
\usepackage{colortbl}
\usepackage{float,wrapfig}
\usepackage{hhline}
\usepackage{multirow}
\usepackage{makecell}
\usepackage{tikz}

\usepackage{url}

\usepackage{algorithm, algorithmic}
\usepackage{enumitem}



%% file: sec/0_abstract.tex
\begin{abstract}
Generating long videos using pre-trained video diffusion models, which are typically trained on short clips, presents a significant challenge. Directly applying these models for long-video inference often leads to a notable degradation in visual quality. This paper identifies that this issue primarily stems from two out-of-distribution (O.O.D) problems: frame-level relative position O.O.D and context-length O.O.D. To address these challenges, we propose \textbf{FreeLOC}, a novel training-free, layer-adaptive framework that introduces two core techniques: Video-based Relative Position Re-encoding (VRPR) for frame-level relative position O.O.D,  a multi-granularity strategy that hierarchically re-encodes temporal relative positions to align with the model's pre-trained distribution, and Tiered Sparse Attention (TSA) for context-length O.O.D, which preserves both local detail and long-range dependencies by structuring attention density across different temporal scales. Crucially, we introduce a layer-adaptive probing mechanism that identifies the sensitivity of each transformer layer to these O.O.D issues, allowing for the selective and efficient application of our methods. Extensive experiments demonstrate that our approach significantly outperforms existing training-free methods, achieving state-of-the-art results in both temporal consistency and visual quality. Code is available at \href{https://github.com/Westlake-AGI-Lab/FreeLOC}{https://github.com/Westlake-AGI-Lab/FreeLOC}.

\end{abstract}

%% file: sec/1_intro.tex
\section{Introduction}
\label{sec:intro}

Video generation has been established as a core research problem in computer vision and generative modeling due to its utility in content creation, simulation, education, and interactive media. In recent years, extensive research has
been conducted in this area, where video diffusion models~\cite{kong2024hunyuanvideo,yang2024cogvideox,wan2025wan,wang2025lavie,guo2023animatediff,chen2024videocrafter2, genmo2024mochi,hacohen2024ltx, blattmann2023stable} have made remarkable progress. Benefiting from training on a large collection of annotated video data, these models can generate high-quality videos that are nearly indistinguishable from reality. 

Despite impressive generation quality, most video diffusion models are designed for short clips. Directly generating longer videos causes significant degradation due to the discrepancy between training and inference lengths. While retraining models is computationally prohibitive, existing training-based autoregressive (AR) methods~\cite{teng2025magi, huang2025self, chen2025skyreels, zhang2025packing, henschel2025streamingt2v,yin2025slow} still struggle to match the quality of native short video generation, posing an open challenge.

Recently, increasing attention has been directed toward training-free approaches that adapt an off-the-shelf short video diffusion model to a long video generator at inference time.
A popular line of work~\cite{qiu2023freenoise, cai2025ditctrl, wang2023gen} partitions the target video into multiple temporally overlapping clips using a sliding window, and processes these clips in an asynchronous or tiled fashion.  
While such approaches effectively maintain consistency within localized video segments, the inherent constraints of the sliding window paradigm restrict interactions between distant frames and consequently impair global temporal coherence. 
Other works~\cite{lu2024freelong,lu2025freelong++,li2025longdiff} attempt to generate long videos in a single pass, and video quality is subsequently refined by manipulating latents to balance spatial detail against temporal consistency. 
 However, these empirical designs are still suboptimal and may still produce artifacts such as identity drift, inconsistent lighting, or abrupt scene transitions.
Moreover, most existing training-free pipelines are built upon UNet-based diffusion architectures, making them incompatible with state-of-the-art DiT-based models such as Wan~\cite{wan2025wan}.  

In this paper, we focus on training-free long video generation using DiT-based diffusion models and address this challenge from three complementary perspectives. 
Firstly, when extending video diffusion transformers to generate sequences substantially longer than their pre-trained temporal window, the 3D relative positional encodings (RoPE), inevitably fall outside the distribution encountered during training. This frame-level relative position out-of-distribution (O.O.D) phenomenon causes the model to misinterpret temporal dependencies, leading to degraded quality.
This issue mirrors the extrapolation problem in RoPE-based LLMs~\cite{han2023lm,jin2024llm,chen2023extending}.
However, naive solutions from LLMs like clipping~\cite{han2023lm} or grouping~\cite{jin2024llm} fail for video because they overlook the hierarchical nature of video's temporal dependencies: nearby frames require fine-grained precision for local dynamics, whereas distant frames contribute to global coherence. To address this, we introduce \textbf{Video-based Relative Position Re-encoding (VRPR)}, a multi-granularity scheme that adaptively remaps out-of-range frame-level relative positions back into the model's familiar domain with varying precision, preserving both short-range fidelity and long-range consistency.

While VRPR alleviates the positional extrapolation issue, generating significantly longer videos can induce another O.O.D problem that we refer to as context-length O.O.D problem. As the temporal sequence grows, the expansion of attention tokens leads to overly diffused attention distributions and increased attention entropy, weakening the model’s ability to focus on informative local regions. Existing methods, such as fixed sliding-window attention~\cite{lu2024freelong, qiu2023freenoise, li2025longdiff}, effectively constrain the receptive field but fail to capture long-range temporal dependencies. To address this trade-off, following the application of VRPR, we introduce \textbf{Tiered Sparse Attention (TSA)}, a hierarchical attention mechanism that adaptively allocates attention density across temporal scales. TSA preserves dense attention within short temporal neighborhoods to maintain fine-grained visual details, while employing progressively narrower striped attention patterns for more distant ones to sustain global coherence. This tiered design ensures that the model allocates sufficient attention to both local motion cues and long-range dependencies without overwhelming the attention space.

Although VRPR and TSA effectively alleviate frame-level relative position and context-length O.O.D issues respectively, applying them uniformly across all transformer layers neglects the heterogeneous roles and sensitivities of individual layers in video DiTs. To better understand this issue, we conduct a detailed empirical study to quantify each layer’s sensitivity to these O.O.D effects. Specifically,  we develop an automatic probing mechanism that quantifies the sensitivity of each DiT's self-attention layer to the two O.O.D sources. Building upon this, we introduce a training-\textbf{free} \textbf{l}ayer-adaptive \textbf{O}.O.D \textbf{c}orrection (\textbf{FreeLOC}) for long video generation, which selectively applies VRPR and TSA to the most responsive layers. For frame-level relative position sensitive layers, we utilize aforementioned VRPR to mitigate frame-level relative position O.O.D problems. For context-length sensitive layers, we employ TSA to constrain the context length following VRPR. This targeted adaptation ensures that positional and contextual corrections are applied precisely where they are most effective, enhancing long-video stability and generation quality.

To validate the effectiveness of FreeLOC, we conduct extensive experiments. The results demonstrate that FreeLOC substantially outperforms existing training-free baselines in both visual fidelity and temporal consistency.
In summary, this paper makes the following contributions:
\begin{itemize}
\item We propose a Video-based Relative Position Re-encoding (VRPR) strategy that remaps frame-level relative positions into the pre-trained range, effectively mitigating frame-level relative-position O.O.D issues.
\item We introduce a Tiered Sparse Attention (TSA) mechanism that preserves the attention distribution under extended context lengths, maintaining a balance between local detail and global temporal coherence.
\item We design a layer-adaptive strategy named FreeLOC which guided by automatic layer sensitivity probing, enabling selective and efficient application of our methods for optimal performance.
\item Extensive experiments verify that FreeLOC achieves state-of-the-art results among training-free approaches for long video generation.
\end{itemize}

%% file: sec/2_related_work.tex
\section{Related Work}
\label{sec:realted_work}
\noindent\textbf{Video Diffusion Model.} 
Recently, significant progress has been made in video diffusion models~\cite{kong2024hunyuanvideo,yang2024cogvideox,wan2025wan,wang2025lavie,guo2023animatediff,chen2024videocrafter2, genmo2024mochi,hacohen2024ltx, blattmann2023stable, Worldforge}, allowing for the synthesis of high-resolution videos that maintain coherent visual and motion continuity over dozens of frames. Early approaches adapted pre-trained Unet-based text-to-image (T2I) diffusion models~\cite{rombach2022high}. These models were enhanced with temporal attention mechanisms to capture the dependencies between video frames. 
A major milestone in generative modeling was the emergence of Diffusion Transformers (DiT)~\cite{peebles2023scalable}, which offer strong scalability and performance. Building on this architecture, recent state-of-the-art video diffusion models such as CogVideoX~\cite{yang2024cogvideox}, Mochi~\cite{genmo2024mochi}, HunyuanVideo~\cite{kong2024hunyuanvideo}, LTX-Video~\cite{hacohen2024ltx}, MovieGen~\cite{polyak2024movie}, and Wan~\cite{wan2025wan} have achieved impressive results.
However, these video DiT models remain limited to short durations, as they are trained only on short clips.

\noindent\textbf{Long Video Generation.}
The significant computational demands associated with training DiT-based architectures have largely confined the output video to a brief duration around 5 seconds. In response to this limitation, several training-based approaches~\cite{teng2025magi, huang2025self, chen2025skyreels, zhang2025packing, henschel2025streamingt2v,yin2025slow, yang2025longlive, cui2025self, liu2025rolling, li2025stable} have adopted an autoregressive mechanism to produce longer videos. 
In addition to these improvements, training-free approaches~\cite{lu2024freelong, qiu2023freenoise, tan2025freepca, zhao2025riflex, kim2024fifo, cai2025ditctrl,lu2025freelong++, li2025longdiff, wang2023gen} for adapting pre-trained short video models to generate long-duration videos have attracted increasing interest. For instance, some methods~\cite{qiu2023freenoise, cai2025ditctrl, wang2023gen} extend video length by integrating overlapping segments through a sliding-window technique.
Other methods~\cite{lu2024freelong, lu2025freelong++, tan2025freepca} explore to identify key components of videos through principal component analysis or frequency analysis during generation and improve output quality by progressively mixing these components.
Closely related to our work, RIFLEx~\cite{zhao2025riflex} and LongDiff~\cite{li2025longdiff} modify positional embeddings to mitigate frame repetition and quality degradation.
However, RIFLEx only supports 2× length extension, while LongDiff, designed for U-Nets, relies on a heuristic mapping that requires 16× attention recomputation, making it impractical for Video DiTs.
More importantly, both methods apply uniform modifications across all layers, ignoring their heterogeneous roles.

%% file: sec/3_method.tex
\section{Method}
\input{figure_tex/framework}
As illustrated in Figure~\ref{fig:framework}, we propose FreeLOC, a layer-adaptive, training-free framework to solve the two O.O.D challenges in long video generation with pre-trained DiT models.

\label{sec:method}
\subsection{Preliminaries}
\noindent\textbf{Self-Attention in Video DiT Blocks.} Leading DiT based video generation models, such as Wan~\cite{wan2025wan}, HunyuanVideo~\cite{kong2024hunyuanvideo} and CogvideoX~\cite{yang2024cogvideox}, leverage a full self-attention~\cite{vaswani2017attention} Transformer framework to effectively model spatiotemporal dependencies within video sequences. 
Such mechanism is formally computed as follows:
\begin{equation}
\small
\text{Attention}(\bm{Q}, \bm{K}, \bm{V}) = \text{softmax}\left(\frac{\text{RoPE}(\bm{Q}) \text{RoPE}(\bm{K})^\top}{\sqrt{d}}\right)\bm{V}, \label{eq:attn}
\end{equation}
where $\bm{Q}, \bm{K}, \bm{V}$ are query, key, value tokens obtained by applying three separate projection matrices to the video tokens, and $d$ is the key dimensionality. 

RoPE in Equation~\ref{eq:attn} injects relative spatial and temporal positional information directly into the query and key tokens before the computation of attention score. For long video generation, only the relative positional information along the temporal dimension needs to be considered. Specifically, for a query frame at index $i$ and a key frame at index $j$, their relative position is given by $P_\text{ori} = i-j$. For a video sequence of length $L$, where both query and key frames are assigned indices from $0$ to $L-1$, the comprehensive extent of all possible relative positions $p$ is delimited by the interval $[-(L-1), L-1]$.

\subsection{Frame-level relative-position O.O.D}

State-of-the-art video DiT models rely on 3D-RoPE for modeling spatio-temporal relations. However, when generating videos longer than the pre-trained duration, frame-level relative positions fall outside the training distribution. This frame-level relative-position O.O.D forces the model to extrapolate RoPE~\cite{zhao2025riflex, chen2023extending}, leading to degraded quality and temporal inconsistency.
This problem parallels the RoPE extrapolation issue in LLMs~\cite{han2023lm,jin2024llm,chen2023extending}, where extended sequence lengths mis-align positional relationships. Prior fixes such as clipping~\cite{han2023lm} or grouping~\cite{jin2024llm} restrict positions to the pre-trained range, but they fail to capture the hierarchical temporal structure of videos, resulting in over-smoothed motions or loss of long-range coherence.

To address this, we introduce \textbf{Video-based Relative Position Re-encoding (VRPR)}, a multi-granularity strategy that  remaps O.O.D relative positions back into the pre-trained range. The design of VRPR is motivated by the empirical observation that the magnitude of attention in video DiTs decays as the temporal distance between frames increases~\cite{li2025radial}. This suggests that nearby frames require high-precision positional information for detail, while distant frames contribute more to global coherence and can be represented with lower precision.
Building on this, VRPR employs a multi-granularity re-encoding scheme that adjusts the precision of relative position information based on frame distance $|i-j|$ between frame $i$ and frame $j$.
As illustrated in the green block of Figure~\ref{fig:framework} (with more details provided in the supplementary), the re-encoding process is structured into three distinct granularities:

\noindent\textbf{Fine-Grained Re-Encoding.} For short-range dependencies within a local window $W_1$ ($ |i-j|\leq W_1$), the original relative positions are retained to fully exploit high-frequency temporal information essential for motion continuity and detail preservation.

\noindent\textbf{Medium-Grained Re-Encoding:} For frames at moderate temporal distances ($W_1 < |i-j| \leq W_2$),  where attention is weaker, we apply a quantized re-encoding. This reduces positional resolution while maintaining the relative order. The original relative position $P_{\text{ori}}=i-j$ is mapped to a new position $P$. Specifically, a \texttt{FLOOR} ($\lfloor \cdot \rfloor$) operation is applied to their relative positions. This operation groups several frames together, treating them as being at the same relative distance. We use a group size of $G_1$ to quantize every $G_1$ frames to the same relative position, formulated as follows:
\begin{equation}
    P = \left\lfloor \frac{P_{\text{ori}}}{G_1} \right\rfloor + \frac{P_\text{ori}}{|P_\text{ori}|}(W_1 - \left\lfloor \frac{W_1}{G_1} \right\rfloor),
    \small
\end{equation}
where $|\cdot|$ denotes the absolute value symbol. This quantization compresses mid-range positions into a coarser granularity, aligning them with the pre-trained range while maintaining the relative order.  
Such representation facilitates a smooth transition of relative positions from the fine-grained to the medium-grained re-encoding area, e.g., relative position $W_1-1$ transitions smoothly to $W_1$. This coarser re-encoding aids in managing the relationships between more distant tokens while retaining a moderate level of positional detail, thereby balancing visual details with temporal consistency.

\noindent\textbf{Coarse-Grained Re-Encoding.} 
For frames at long temporal distances ($|i-j| > W_2$),  the attention primarily serves to maintain global coherence, we apply a more aggressive quantization with a larger group size $G_2$. This strongly compresses the positional range of distant frames, retaining only their approximate ordering, and the formal formulation is as follows:

\begin{equation}
    P = \left\lfloor \frac{P_{\text{ori}} }{G_2} \right\rfloor + \frac{P_\text{ori}}{|P_\text{ori}|}(W_2 - \left\lfloor \frac{W_2}{G_2} \right\rfloor - \left\lfloor \frac{W_2 - W_1}{G_1} \right\rfloor),
    \small
\end{equation}
This operation compresses the positional range of distant frames, effectively mapping them back into the pre-trained positional domain while retaining their approximate ordering. This coarse quantization preserves global temporal alignment without causing perceptual drift.

By employing aforementioned multi-level re-encoding, the VRPR framework effectively constrains the frame-level relative positions to the pre-trained range of the video DiT. This approach, which aligns with the inherent characteristics of attention decay, allows for the generation of long videos that better capture local details while preserving global consistency.

\subsection{Context-length O.O.D}

Even with VRPR mapping relative positions back into the pre-trained range, generating very long videos still causes blurred details. This occurs because extremely long token sequences expand the softmax domain, making attention weights overly diffuse and weakening fine-grained motion cues. Recent studies in LLMs~\cite{han2023lm} and image models~\cite{jin2023training} show that such attention diffusing manifests as increased attention entropy, which correlates with degraded generation quality. The same effect appears in video (with details shown in supplementary), which we term context-length O.O.D. 
A common remedy is fixed-length sliding-window attention~\cite{lu2024freelong, qiu2023freenoise, li2025longdiff}, which preserves local details but breaks long-range temporal dependencies, harming global video consistency.

To address this, we propose \textbf{Tiered Sparse Attention (TSA)}, a hierarchical attention mechanism designed to restrict the effective context length within the pre-trained range while preserving long-term consistency.  
TSA operates at multiple temporal granularities, analogous to the multi-level hierarchy in VRPR, and incorporates an \textit{attention sink} that serves as a persistent global anchor. Here, we still define temporal distance between two frame $i$ and $j$ as $|i-j|$.
A visualization of the mask structure is provided in blue area of Figure~\ref{fig:framework}. 

\noindent\textbf{Local Window Attention.}
For short-term dependencies (temporal distance $|i-j| \le D_1$), a standard, dense self-attention window $D_1$ is used. This ensures that fine details and rapid, local motions are captured with high fidelity.

\noindent\textbf{Striped Attention for Mid-Range Dependencies.}
For medium-range dependencies ($D_1 < |i-j| \le D_2$), TSA introduces a \textit{striped attention pattern}. 
 This design is motivated by the empirical observation in video DiTs~\cite{li2025radial} that beyond local spatial space, high attention scores often occur between tokens at corresponding spatial locations across frames, which is crucial for temporal coherence.
Specifically, as illustrated in TSA part of Figure~\ref{fig:framework}, our striped attention utilize shrunk window size $D_s$ to enable a token at a specific spatial position to interact with tokens in the similar spatial locations across frames. This striped approach effectively reduces the context length while simultaneously expanding the temporal receptive field to enhance temporal consistency.
While local window attention captures rapid motion, striped attention aggregates context to ensure subject and scene consistency, striking a balance between detail preservation and temporal consistency.

\noindent\textbf{Long-Range Pruning \& Attention Sink.}
For long-range interactions ($|i-j| > D_2$), direct attention is pruned. This decision is based on our subsequent empirical finding that interactions with overly distant tokens can lead to a degradation of details. To ensure global consistency, TSA designates the initial frame as an \textbf{attention sink}. Tokens from all subsequent frames are permitted to attend to this first frame, which acts as a persistent global anchor, enforcing long-term coherence throughout the entire sequence.

Specifically, we construct a 4D attention mask $\tilde M \in \{1, 0\}^{f \times f \times n \times n}$ shown in the TSA part of Figure~\ref{fig:framework} , where $f$ denotes the frame number and $n$ denotes the number of tokens per frame. Here, each element $\tilde M_{i,j,k,l} = 1$ indicates that the token at spatial  position $k$ in frame $i$ is permitted to attend to the token at position $l$ in frame $j$. Conversely, $\tilde M_{i,j,k,l} = 0$ denotes that attention between the token pair is suppressed. The aforementioned TSA attention mask is constructed as follows:
 \begin{equation} 
\tilde M_{i,j,k,l} = \begin{cases} 
    1, & \text{if } |i-j| < D_1 \text{ or } j=0 \\
    1, & \text{if } D_1 \le |i-j| < D_2 \text{ and } |k-l| < D_\text{s} \\
    0, & \text{otherwise}
\end{cases}
\small
\end{equation}

Specifically, $D_s= \lfloor\frac{nD_1}{\alpha(D_2-D_1)}\rfloor$. This implies that the actual token computation is reduced by a factor of $\alpha$ compared to the standard window size. By limiting the total number of tokens participating in the attention mechanism, TSA avoids the dilution of attention weights, which enhances the preservation of both visual and motion details. More importantly, this design expands the temporal receptive field for interactions, enabling better temporal consistency compared to the conventional sliding window attention mechanism. Consequently, TSA achieves a superior balance between temporal coherence and the fine-grained details in the generated video.

\subsection{Layer-wise Probing Procedure for Two O.O.D Issues}
While VRPR and TSA address two O.O.D issues, applying them uniformly across all layers ignores the heterogeneous sensitivities within video DiTs. Some layers are far more affected by positional shifts, while others are more vulnerable to long-context expansion, making uniform application suboptimal.
To characterize these differences, we conduct a {layer-wise probing analysis} on Wan2.1-T2V-1.3B~\cite{wan2025wan}. The results show that \textit{self-attention layers exhibit markedly different sensitivities to the two O.O.D sources}. The subsections below describe the probing design and sensitivity metrics for each case.

\noindent\textbf{Probing Layer-wise Sensitivity to Frame-level Relative Position O.O.D}.
To study this O.O.D issue, we design an automated probing procedure that quantifies the importance of relative position information in each self-attention layer. We first generate $M$ diverse prompts and synthesize $N$ original videos for each, using the default pre-trained length (81 frames for Wan2.1-T2V-1.3B).
We then create a corresponding set of probing videos, generated with the same length but with a controlled perturbation applied to only \textbf{one} self-attention layer at a time. The perturbation modifies the RoPE only for keys $\mathbf{K}$ while keeping queries $\mathbf{Q}$ unchanged (equivalent to shifting $\mathbf{Q}$ alone). This selectively alters frame-level relative positions in that layer, simulating values beyond the pre-trained range.
We evaluate four frame-level shifts, $\pm20$ and $\pm40$. For example, a $+20$ shift moves the default position index of $\mathbf{K}$ from $[0,20]$ to $[20,40]$, changing the relative position interval from the trained $[-20,20]$ to an O.O.D range $[-40,0]$.

Following the acquisition of all $4\times M\times N$ sets of original videos and probing videos, we employ two metrics to quantify each layer's sensitivity. First, we measure the average Vision Reward~\cite{xu2024visionreward} (higher is better) for each probing videos which approximates the human rating of the generated probing videos' quality after shifting frame-level position.
A lower Vision Reward score signifies that the layer is sensitive to frame-level relative position O.O.D, as generation quality deteriorates when encountering untrained relative positions.

Moreover, it is observed in LLMs' extrapolation~\cite{han2023lm, jin2024llm} that relative positions O.O.D in RoPE can induce significant fluctuations in the distribution of attention logits.  We compute the average attention logits $w_i$ across all heads and sampling steps for each layer $i$. Here we measure the \textbf{A}ttention \textbf{L}ogits \textbf{D}ifference (ALD) as the metric which is defined as  the relative norm of the difference between the logits of the probing and original videos:$\|w_i^\text{probing}-w_i^\text{original}\| \ / \ |w_i^\text{original}\|$. A higher ALD indicates a substantial change in the attention mechanism's behavior, signifying high sensitivity to positional O.O.D. The statistical results are shown in Figure~\ref{fig:attn_logits_visonreward}. Moreover, Figure~\ref{fig:rope_probe_quality} illustrates the visualization outcomes for Layer 18 which is most insensitive to frame-level relative position O.O.D and Layer 28 which is most sensitive to frame-level relative position O.O.D.
\input{figure_tex/attn_logtis_and_visionreward}

\input{figure_tex/rope_probing_quality_res}

\noindent\textbf{Probing Layer-wise Sensitivity to Context Length O.O.D}.
Following the aforementioned probing approach, we also employ $M$ diverse textual descriptions for video generation and each generate $N$ videos leveraging different random seed.
Specifically, to probe layer-wise sensitivity to context length O.O.D, we first generate original videos by extending the sequence length longer than pre-trained length while restricting the frame-level relative positions within the pre-trained range utilizing VRPR. Subsequently, building upon this configuration, we generate the probing video by employing a layer-by-layer conventional sliding window attention mechanism. This approach confines the context length of each layer to the pre-trained range, while the self-attention configurations of the other layers remain unchanged. For the probing results presented below, we employed a simple yet generalizable pre-training length extension factor of $2\times$, as similar patterns were observed with other scaling factors (like $3\times$, $4\times$).

To quantify the impact of the extended context, we leverage attention entropy as a diagnostic tool. Attention entropy measures the uniformity of the attention distribution within the self-attention mechanism~\cite{han2023lm, jin2023training}, and it has been observed that attention entropy tends to increase as the context length grows because of the dispersed attention. 
Following this, we define a Context Length Sensitivity Score, $S_i$, for each layer $i$. This score is formulated as the relative difference between the attention entropy $H_i^{\text{original}}$  computed in layer $i$'s self-attention in the original video and the attention entropy $H_i^\text{probing}$ computed in layer $i$'s self-attention in the probing video. The formulation is as follows:
\begin{equation}
S_{i} = \frac{\|H_i^{\text{probing}} - H_i^\text{original}\|}{\|H_i^\text{original}\|}, \label{eq:score_entropy}
\small
\end{equation}
This metric provides a normalized measure of how significantly the attention distribution at a specific layer is perturbed by the increase in context length. A low $S_i$ indicates that the layer's attention mechanism maintains a similar distribution regardless of whether it processes the full extended context or a constrained one, thereby demonstrating insensitivity to context length O.O.D.
Conversely, a high value of $S_i$ indicates significant variation in the layer’s attention distribution with changing context length, suggesting sensitivity to context-length O.O.D. 
The probing result is shown in Figure~\ref{fig:attn_entropy}.

\input{figure_tex/attention_entropy}

\subsection{FreeLOC: Layer-Adaptive O.O.D Correction Strategies for Long Video Generation}
Building upon the observation that individual layers within the model exhibit varying degrees of sensitivity to two O.O.D issues, we introduce \textbf{FreeLOC}, a layer-adaptive strategy  that applies VRPR and TSA selectively (The specific sensitivity profile for each layer is detailed in the supplementary).
For layers sensitive only to frame-level positional O.O.D, we apply \textbf{VRPR} to re-encode relative positions while preserving full token interactions, thereby maintaining temporal consistency. 
For layers sensitive to context-length O.O.D, regardless of their positional sensitivity, we apply a two-step scheme: first \textbf{VRPR} for positional correction, then \textbf{TSA} to restrict effective context. This combination mitigates both O.O.D. issues and preserves fine-grained spatial and motion details. 

%% file: figure_tex/framework.tex
\begin{figure*}[t]
  \centering
  \vspace{-10pt}
  \includegraphics[width=1\linewidth]{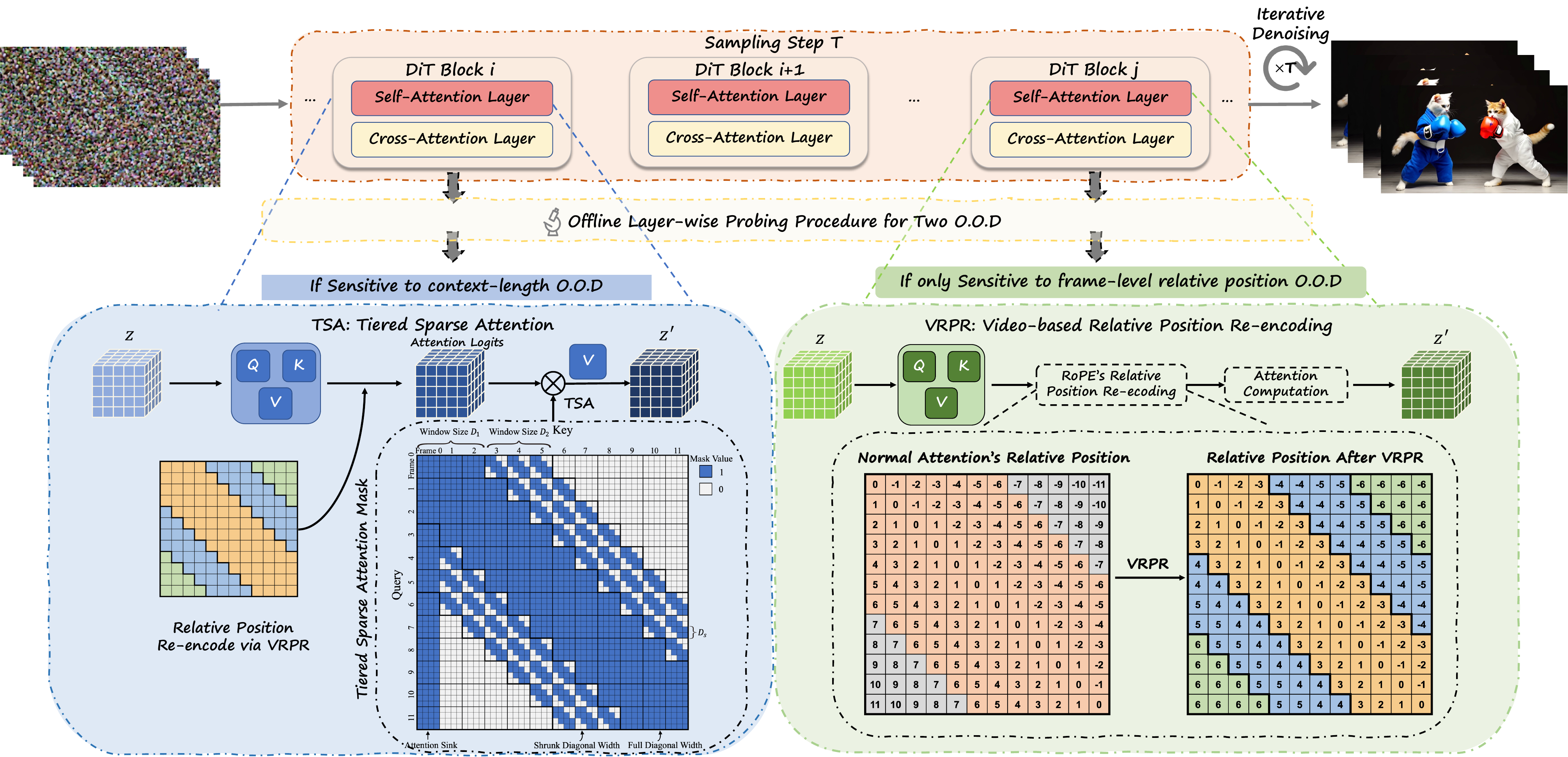}
   \caption{\textbf{Overview of FreeLOC.} FreeLOC employs an offline, layer-wise probing procedure to identify DiT layers sensitive to two O.O.D sources: context-length O.O.D and frame-level relative position O.O.D. For layers only sensitive to frame-level relative position, we apply VRPR strategy to hierarchically remap out-of-range positions back into the pre-trained domain.  For layers sensitive to context-length, we utilize TSA combined with VRPR to balance local detail and global coherence.  Zoom in for better view.}
   \label{fig:framework}
   \vspace{-10pt}
\end{figure*}

%% file: figure_tex/attn_logtis_and_visionreward.tex
\begin{figure}[t]
  \centering
  \includegraphics[width=1\linewidth]{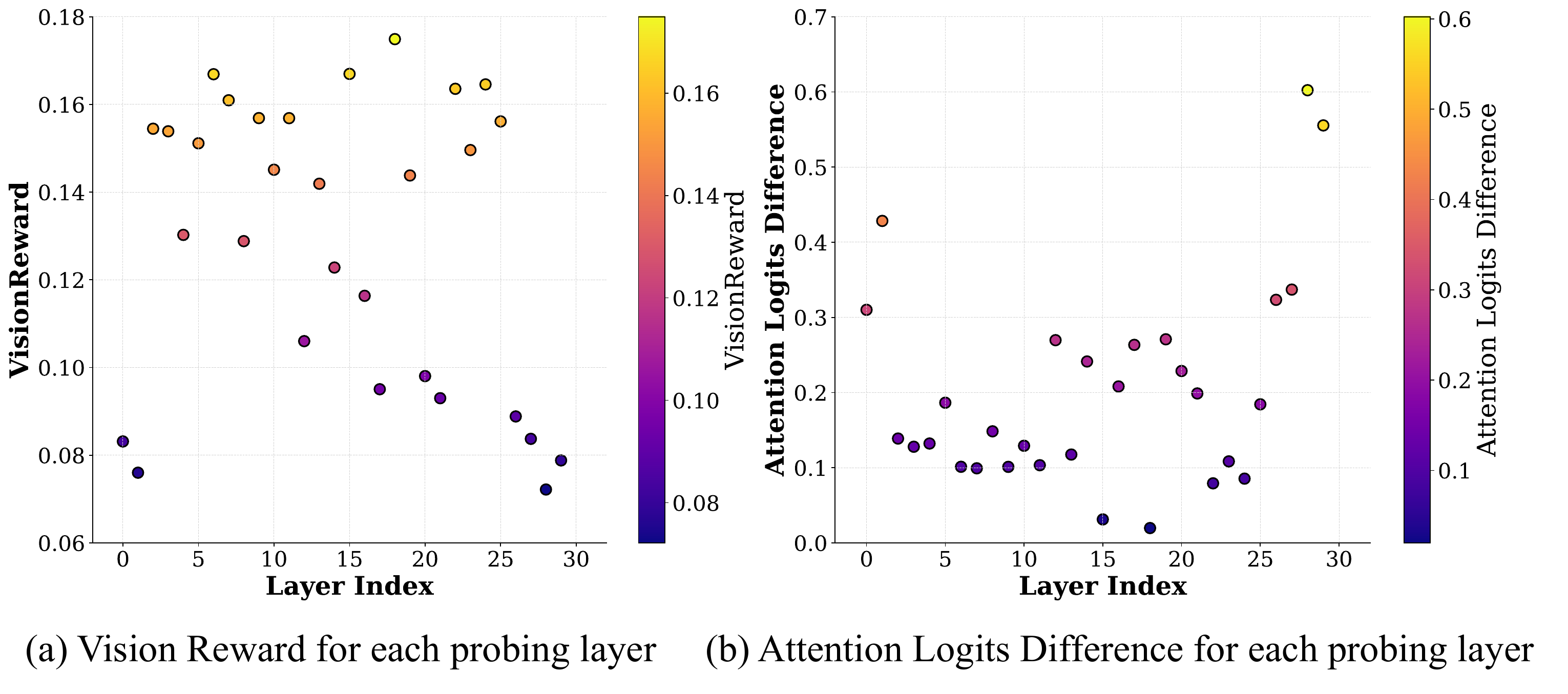}
   \caption{\textbf{Sensitivity analysis for each layer during frame-level relative position O.O.D probing. }
        (a) Vision Reward for each probing layer. A lower score indicates that the layer is more sensitive to frame-level relative position O.O.D.
        (b) Attention Logits Difference (ALD) for each probing layer. A higher value indicates a significant change in the behavior of the attention mechanism, signifying high sensitivity to positional O.O.D. }
   \label{fig:attn_logits_visonreward}
   \vspace{-5pt}
\end{figure}

%% file: figure_tex/rope_probing_quality_res.tex
\begin{figure}[t]
\vspace{-8pt}
  \centering
  \includegraphics[width=1\linewidth]{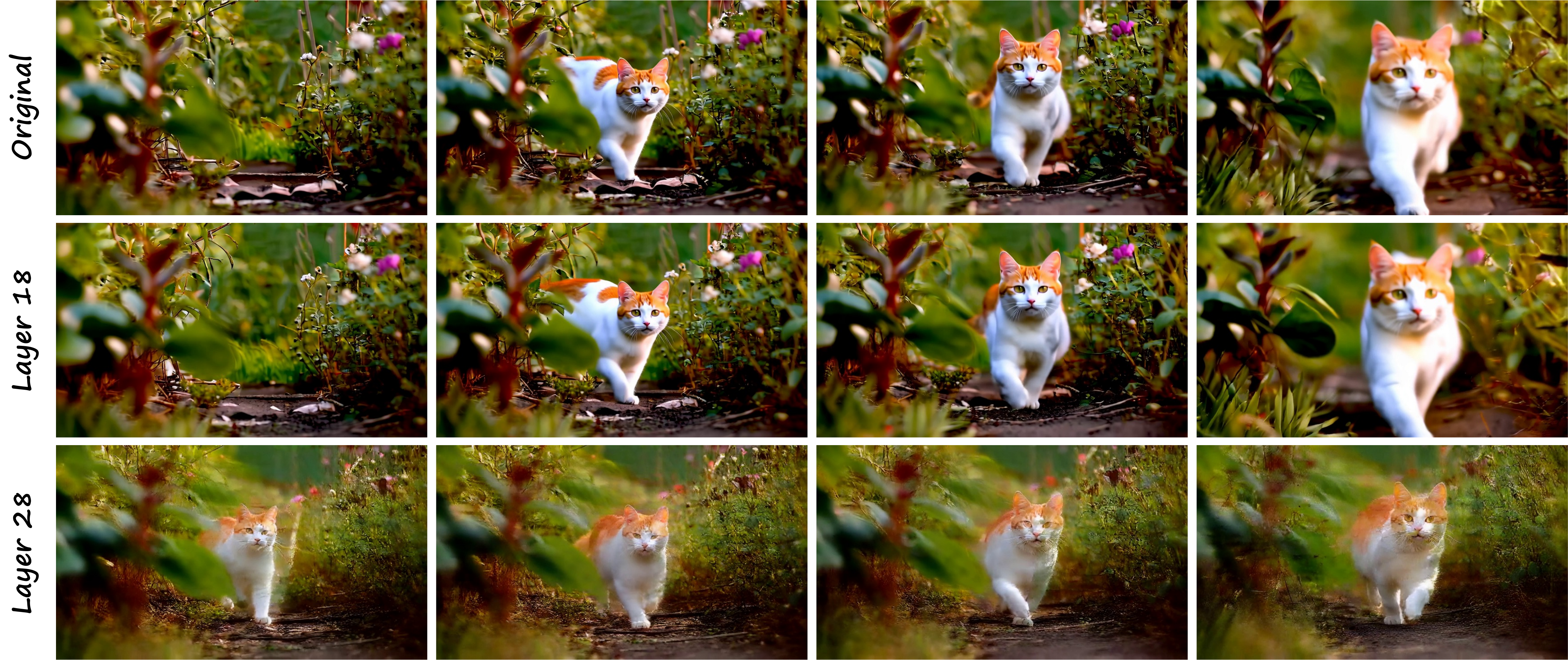}
   \caption{
   \textbf{Layer-wise sensitivity to frame-level relative-position O.O.D.}
The figure compares the original video with probing outputs from Layer 18 (low sensitivity) and Layer 28 (high sensitivity). Stronger sensitivity leads to noticeable distortions when relative-position shifts are applied.
   }
   \label{fig:rope_probe_quality}
   \vspace{-10pt}
\end{figure}

%% file: figure_tex/attention_entropy.tex
\begin{figure}[t]
  \centering
  \includegraphics[width=0.65\linewidth]{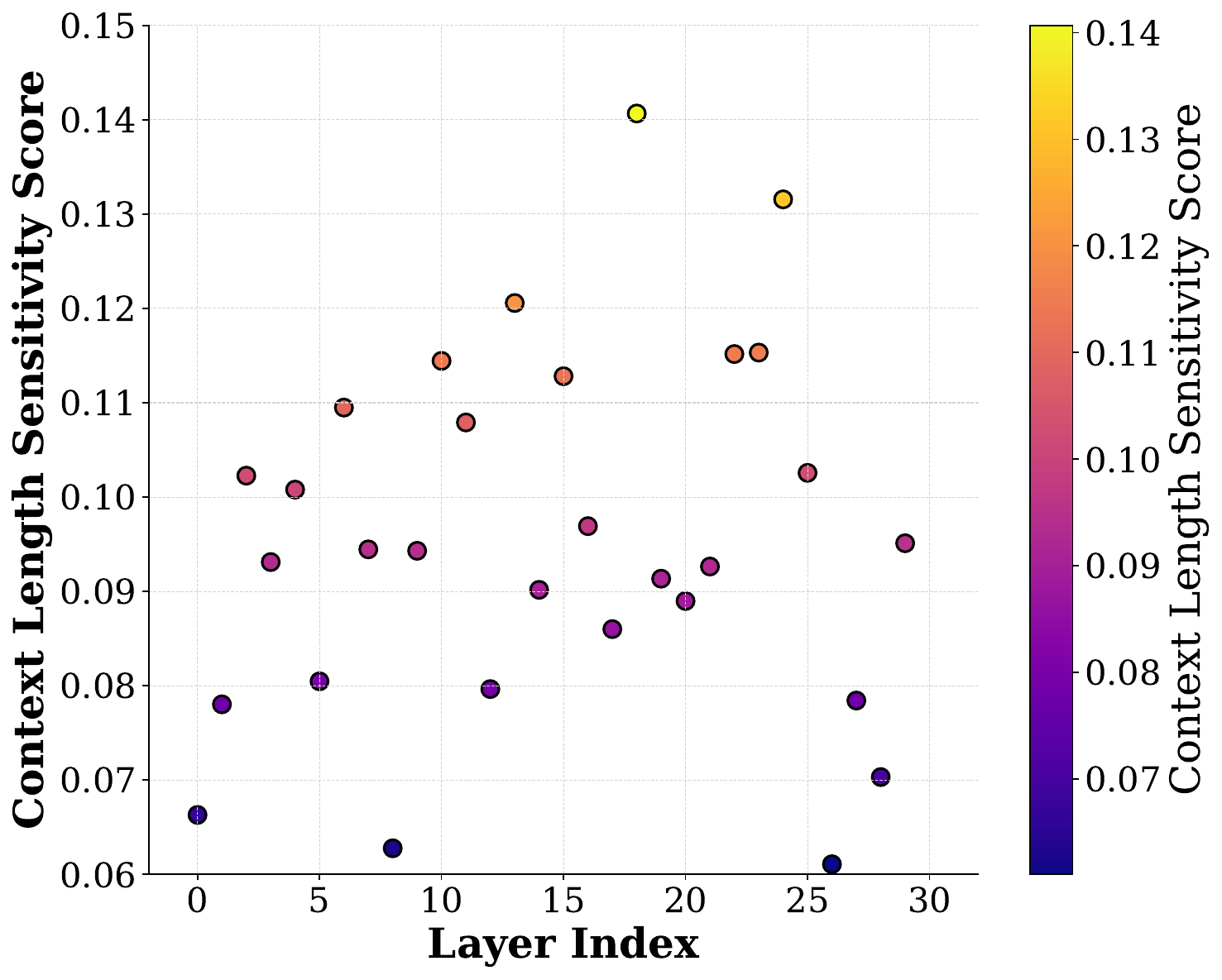}
  \vspace{-8pt}
   \caption{
   \textbf{Layer-wise sensitivity to context-length O.O.D measured via attention entropy differences.}
   }
   \label{fig:attn_entropy}
   \vspace{-15pt}
\end{figure}

%% file: sec/4_experiments.tex
\section{Experiments}
\label{sec:experiments}

\input{table_tex/comparison}

\subsection{Implementation Details}
\noindent\textbf{Setting up.} To validate the broad applicability of FreeLOC, we integrated our methods with two publicly accessible video DiT models: Wan2.1-T2V-1.3B~\cite{wan2025wan} and HunyuanVideo~\cite{kong2024hunyuanvideo} in a training-free manner. For experiments, we generate videos at $480$p ($832\times480$) with $2\times$ and $4\times$ length extension.

\noindent\textbf{Evaluation metrics.} Following prior work~\cite{lu2024freelong, lu2025freelong++, tan2025freepca, zhao2025riflex}, we employ metrics from Vbench~\cite{huang2024vbench} to assess FreeLOC. We evaluate from two aspects: video consistency and video quality. Specifically, for video consistency, we utilize following metrics:
(1) Subject consistency (SC). (2) Background consistency (BC). (3) Motion Smoothness (MS). For video quality, we use following metrics: (1) Imaging Quality (IQ). (2) Aesthetic Quality (AQ). (3) Dynamic Degree (DD).

\noindent\textbf{Baseline.} A comparative evaluation is conducted between the proposed method and several existing training-free technique for long video generation using diffusion models. The benchmarked approaches include: (1) Direct sampling; (2) Sliding window, which employs temporal sliding windows to process video content in fixed-length frame segments; (3) FreeNoise~\cite{qiu2023freenoise}, a method that improves long-range temporal consistency through the reuse of input noise patterns; (4) Freelong~\cite{lu2024freelong}, a method that integrates low-frequency global features with high-frequency local attention maps to enhance video quality; (5) RIFLEx, an approach that suppress frame repetition by reducing the intrinsic frequency. For fair comparison, we  implement all baseline methods on the SOTA video DiT models: Wan2.1-T2V-1.3B~\cite{wan2025wan} and Hunyuan Video~\cite{kong2024hunyuanvideo}.

\subsection{Comparison Results}
\noindent\textbf{Quantitative Results.}
Table~\ref{tab:quantitative} reports the quantitative comparison for $2\times$ and $4\times$ extension. 
Direct sampling and RIFLEx suffer from strong degradation, especially on long videos. 
Sliding Window and FreeNoise improve local quality but introduce temporal inconsistency. 
FreeLong yields mild gains but remains limited in both consistency and quality. 
FreeLOC achieves the best overall performance, delivering consistent and high-fidelity long videos.

\noindent\textbf{Qualitative Results.}
Figure~\ref{fig:quality} shows qualitative comparisons on Wan2.1-T2V-1.3B and Hunyuan Video with $4\times$ extension. 
Direct Sampling and RIFLEx exhibit severe loss of visual and motion details. 
Sliding Window and FreeNoise produce sharper frames but break temporal coherence. 
FreeLong offers slight improvements yet still loses fine details and dynamic motion. 
In contrast, our FreeLOC preserves strong temporal consistency while maintaining high visual and motion quality.

\input{table_tex/ablation_layer_wise}

\input{figure_tex/quality_wan}

\subsection{Ablation Studies}
To evaluate the effectiveness of each component within FreeLOC, we conducted a series of detailed ablation studies. For this analysis, we equipped the baseline short video model, Wan2.1-T2V-1.3B, with our proposed methods as an example.

\noindent\textbf{Impact of Layer-wise Strategy and Main Components.}
We first investigate the impact of our layer-wise strategy for applying TSA and VRPR. We compare the performance of the following different configurations: 
(1) Direct Sampling.
(2) Direct+TSA: Applying only TSA across all layers.
(3) Direct+VRPR: Applying only VRPR across all layers.
(4) Direct+$\text{(TSA+VRPR)}_\text{uniform}$: Applying VRPR combined with TSA uniformly across all layers.
(5) Direct+$\text{(TSA+VRPR, VRPR)}_\text{random}$: A random assignment of either VRPR or TSA to each layer.
(6) Direct+$\text{(TSA, VRPR)}_\text{layer-wise}$): Selectively applies VRPR and TSA (instead of VRPR+TSA) to different layers based on their sensitivity to two O.O.D issues.
(7) FreeLOC (Direct+$\text{(TSA+VRPR, VRPR)}_\text{layer-wise}$): Our layer-wise strategy, which selectively applies VRPR and TSA+VRPR to different layers based on their sensitivity to two O.O.D issues.
The experimental results in Table~\ref{tab:ablation_layerwise}, demonstrate that our targeted layer-wise strategy, VRPR and TSA all contribute to the outstanding performance.

\noindent\textbf{Impact of Different Re-encoding Strategies for Frame-level Relative Position O.O.D.}
To validate the effectiveness of our VRPR method in handling frame-level relative positions O.O.D, we compare it against several alternative re-encoding strategies: 
(1) Clipping~\cite{han2023lm}: Truncating relative position embeddings that exceed the range seen during training.
(2) Grouping~\cite{jin2024llm}: Assigning a single, shared positional embedding to groups of adjacent relative positions.
(3) VRPR: Our proposed multi-granularity re-encoding method.
As shown in Table~\ref{tab:ablation_reencoding}, our VRPR approach with three granularity levels effectively addresses the challenges of frame-level relative positions O.O.D, maintaining both temporal consistency and visual details.

\noindent\textbf{Impact of Different Attention Mechanisms for Context-Length O.O.D.}
Finally, we evaluate the contribution of our TSA mechanism for managing context-length O.O.D issues by comparing it with other common attention strategies:
(1) Sliding Window Attention~\cite{lu2024freelong,qiu2023freenoise}.
(2) Selected Frame Attention: An approach proposed by LongDiff~\cite{li2025longdiff}, where a few selected frames are designated as global references for attention computation.
(3) TSA.
The results presented in Table~\ref{tab:ablation_attenton} confirm that our TSA mechanism is more effective at maintaining long-range temporal coherence and visual quality compared to the other attention mechanisms.

\input{table_tex/ablation_reencoding}
\input{table_tex/ablation_attention}

%% file: table_tex/comparison.tex
\begin{table*}[t]
	\centering
    \vspace{-15pt}
	\setlength{\tabcolsep}{3pt}{
    \caption{\textbf{Quantitative results at the extended (2× and 4$\times$) video lengths.} In a training-free manner, our method maintains the quality regarding video consistency and video quality in multiple VBench dimensions  when the generation length grows.}
    \label{tab:quantitative}
    \footnotesize
        \begin{tabular}{lccccccccc}
            \toprule
            \multirow{3}{*}{\textbf{Model}} & \multirow{3}{*}{\#Frames} & \multirow{3}{*}{Method}  &\multicolumn{3}{c}{Video Consistency}  & \multicolumn{3}{c}{Video Quality} \\
            \cmidrule(lr){7-9}\cmidrule(lr){4-6}
            & & &\makecell{Subject\\Consistency($\uparrow$)}&\makecell{Background\\Consistency($\uparrow$)}&\makecell{Motion\\Smoothness($\uparrow$)}&\makecell{Imaging\\Quality($\uparrow$)}& \makecell{Aesthetic\\Quality($\uparrow$)} &\makecell{Dynamic\\Degree($\uparrow$)}\\
            \midrule
            \multirow{12}{*}{\makecell{Wan2.1\\-T2V-1.3B}}
            & \multirow{6}{*}{161 (2$\times$)}
             & Direct Sampling & \underline{97.33}  & \underline{97.23} &  98.69 & 60.34 & 52.72 & 19.31 \\
            & & Sliding Window & 96.48& 96.22 & 98.50 & 64.64 & 54.42 & \textbf{40.82}\\
            & & RIFLEx~\cite{zhao2025riflex} & 97.21 & 97.14 & 98.70 &60.62 & 53.67 & 23.45 \\
            & & FreeLong~\cite{lu2024freelong} & 97.05 & 97.19& \underline{98.84} & 63.91 & 54.56 & 32.19  \\
            & & FreeNoise~\cite{qiu2023freenoise} & 96.82 & 97.10 & 98.82 & \underline{67.19} & \underline{56.01} & 36.34 \\
            & & \cellcolor{mitblue}FreeLOC (Ours) & \cellcolor{mitblue}\textbf{98.06}& \cellcolor{mitblue}\textbf{97.49} & \cellcolor{mitblue}\textbf{98.98} & \cellcolor{mitblue}\textbf{68.31} & \cellcolor{mitblue}\textbf{62.33} & \cellcolor{mitblue}\underline{39.41} \\
            \cmidrule(lr){2-9}
            & \multirow{6}{*}{\makecell{321 (4$\times$)}}
             & Direct Sampling & \textbf{98.50}  & \textbf{97.89} &  98.83  & 59.21 & 49.43 & 4.32  \\
            & & Sliding Window & 96.15 & 95.92 & 98.54 & 65.64 & 54.04 & \textbf{39.81}\\
            & & RIFLEx~\cite{zhao2025riflex} & 98.41 & \underline{97.87} & 98.86 & 59.92 & 49.67 & 4.45 \\
            & & FreeLong~\cite{lu2024freelong} & 97.88 & 97.51 & \underline{98.91} & 63.17 & 54.56 & 21.21  \\
            & & FreeNoise~\cite{qiu2023freenoise} & 97.31 & 97.25 & 98.84 & \underline{66.32} & \underline{56.01} & 35.11 \\
            & & \cellcolor{mitblue}FreeLOC (Ours) & \cellcolor{mitblue}\underline{98.44}& \cellcolor{mitblue}97.78 & \cellcolor{mitblue}\textbf{98.97} & \cellcolor{mitblue}\textbf{67.44} & \cellcolor{mitblue}\textbf{61.21} & \cellcolor{mitblue}\underline{36.27} \\
            \midrule
            \multirow{12}{*}{\makecell{Hunyuan\\Video}}
            & \multirow{6}{*}{253 (2$\times$)}
            & Direct Sampling & 97.24  & 97.18 &  98.78  & 60.79 & 54.43 & 23.32  \\
            & & Sliding Window & 96.35 & 96.01 & 98.67 & 67.02 & \underline{60.07} & \textbf{42.39}\\
            & & RIFLEx~\cite{zhao2025riflex} & \underline{97.32} & \underline{97.29} & \textbf{99.03} & 62.43 &  57.02 & 29.13 \\
            & & FreeLong~\cite{lu2024freelong} & 97.22 & 97.10 & 98.93 & 63.47 & 56.56 & 21.21  \\
            & & FreeNoise~\cite{qiu2023freenoise} & 96.89 & 96.72 & 98.90 & \underline{64.68} & 57.87 & 37.11 \\
            & & \cellcolor{mitblue}FreeLOC (Ours) & \cellcolor{mitblue}\textbf{97.92} & \cellcolor{mitblue}\textbf{97.34} & \cellcolor{mitblue}\underline{98.99} & \cellcolor{mitblue}\textbf{68.92} & \cellcolor{mitblue}\textbf{62.38} & \cellcolor{mitblue}\underline{40.22}  \\
            \cmidrule(lr){2-9}
            & \multirow{6}{*}{509 (4$\times$)}  
            & Direct Sampling & \underline{98.62} & \textbf{98.26} &  98.90  & 58.12 & 51.41  & 6.32  \\
            & & Sliding Window & 96.04 & 95.98 & 98.71 & \underline{66.40} & \underline{59.32} & \underline{38.97}\\
            & & RIFLEx~\cite{zhao2025riflex} & \textbf{98.69} & \underline{98.24} & \textbf{99.01} & 56.56 & 51.67 & 8.45 \\
            & & FreeLong~\cite{lu2024freelong} & 97.88 & 97.51 & 98.91 & 62.83 & 54.56 & 21.21  \\
            & & FreeNoise~\cite{qiu2023freenoise} & 97.31 & 97.25 & 98.88 & 63.91 & 56.01 & 35.11 \\
            & & \cellcolor{mitblue}FreeLOC (Ours) & \cellcolor{mitblue}98.47 & \cellcolor{mitblue}98.12 & \cellcolor{mitblue}\underline{98.95} & \cellcolor{mitblue}\textbf{67.92} & \cellcolor{mitblue} \textbf{61.09} & \cellcolor{mitblue}\textbf{39.28}\\
            \bottomrule
        \end{tabular}
	}
        \vspace{-0.5em}
\end{table*}

%% file: table_tex/ablation_layer_wise.tex
\begin{table}[t]
\centering
    \setlength{\tabcolsep}{1pt}
    \vspace{-5pt}
    \caption{\textbf{Ablation study of layer-wise strategy and main components.}}
    \vspace{-8pt}
    \label{tab:ablation_layerwise}
    \scriptsize \centering
    \begin{tabular}{lccccccc}
    \toprule
      Method  & SC ($\uparrow$) & BC ($\uparrow$) & MS ($\uparrow$) & IQ ($\uparrow$) & AQ ($\uparrow$)& DD ($\uparrow$)\\
    \midrule
    Direct & \textbf{98.50}  & \textbf{97.89} &  98.83  & 59.21 & 49.43 & 4.32  \\
    Direct+TSA & 97.41 & 96.76 & 98.67 & 65.87 & 57.05 & \textbf{37.01} \\
    Direct+VRPR & 98.42 & \underline{97.81} &  98.89 & 61.88 & 54.13 & 15.32 \\
    Direct+$\text{(TSA+VRPR)}_\text{uniform}$ & 97.56 & 97.67 &  98.75 & 65.19  & 56.34 & 34.44 \\
    Direct+$\text{(TSA+VRPR, VRPR)}_\text{random}$& 98.03 & 97.61 &  98.91 & 63.90 & 54.44 & 33.13 \\
    Direct+$\text{(TSA, VRPR)}_\text{layer-wise}$& 98.38 & 97.66 &  98.95 & 66.92 & 60.88 & 35.05 \\
    Direct+$\text{(TSA+VRPR, VRPR)}_\text{layer-wise}$& \cellcolor{mitblue}\underline{98.44}& \cellcolor{mitblue}97.78 & \cellcolor{mitblue}\textbf{98.97} & \cellcolor{mitblue}\textbf{67.44} & \cellcolor{mitblue}\textbf{61.21} & \cellcolor{mitblue}\underline{36.27} \\
    \bottomrule
    \end{tabular}
    \vspace{-8pt}
\end{table}

%% file: figure_tex/quality_wan.tex
\begin{figure*}[t]
  \centering
  \vspace{-5pt}
  \includegraphics[width=1\linewidth]{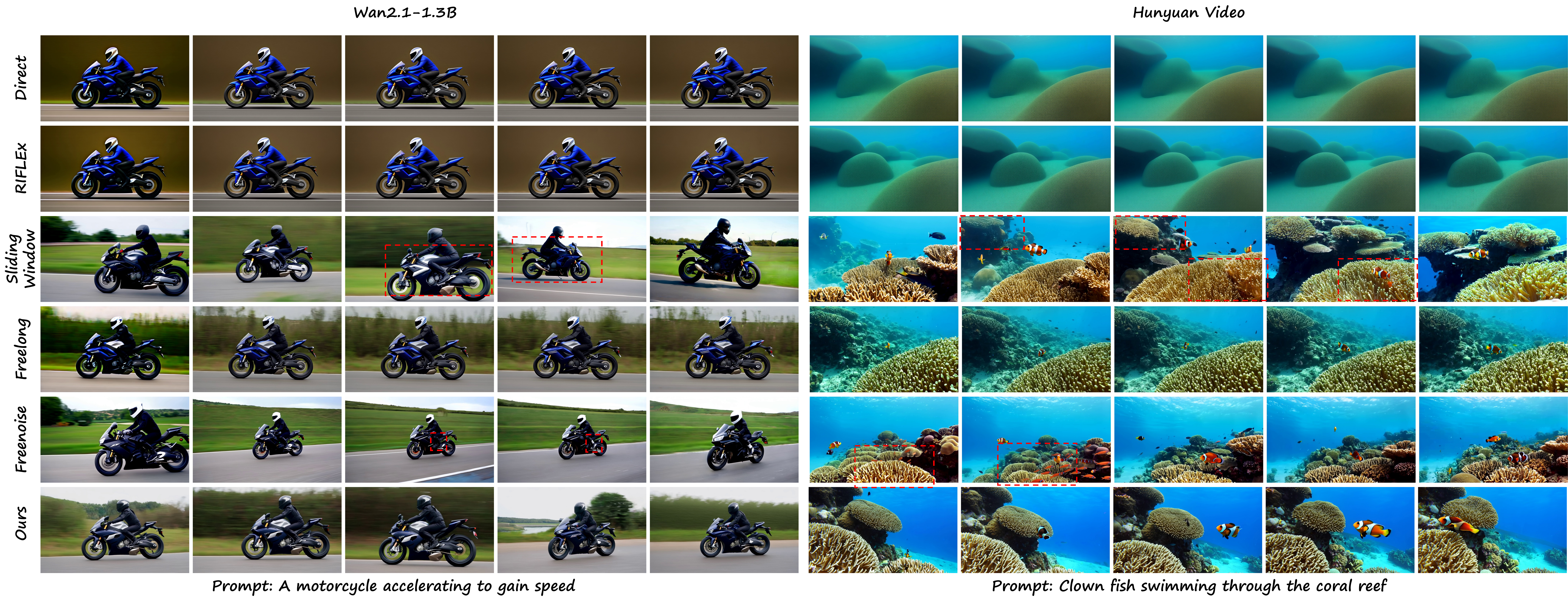}
   \caption{\textbf{Qualitative Results of Wan2.1-1.3B and Hunyuan Video}}
   \label{fig:quality}
   \vspace{-10pt}
\end{figure*}

%% file: table_tex/ablation_reencoding.tex
\begin{table}[t]
\centering
    \vspace{-5pt}
    \setlength{\tabcolsep}{6pt}
    \caption{\textbf{Ablation study of different relative position re-encoding strategy for frame-level relative position O.O.D.}}
    \vspace{-10pt}
    \label{tab:ablation_reencoding}
    \scriptsize \centering
    \begin{tabular}{lccccccc}
    \toprule
      Method  & SC ($\uparrow$) & BC ($\uparrow$) & MS ($\uparrow$) & IQ ($\uparrow$) & AQ ($\uparrow$)& DD ($\uparrow$)\\
    \midrule
    Clipping & 97.83 &97.51 & 98.78 & 60.23 & 52.03& 22.21 \\
    Group & 97.95 & 97.49 &98.66 & 59.91 & 51.32 & 18.32 \\
    Our VRPR& \cellcolor{mitblue}\textbf{98.44
    }& \cellcolor{mitblue}\textbf{97.78} & \cellcolor{mitblue}\textbf{98.97} & \cellcolor{mitblue}\textbf{68.84} & \cellcolor{mitblue}\textbf{61.21} & \cellcolor{mitblue}\textbf{36.27} \\
    \bottomrule
    \end{tabular}
    \vspace{-10pt}
 \end{table}

%% file: table_tex/ablation_attention.tex
\begin{table}[t]
    \centering
    \setlength{\tabcolsep}{3pt}
    \vspace{-2pt}
    \caption{\textbf{Ablation study of different attention mechanism for context-length O.O.D.}}
    \vspace{-6pt}
    \label{tab:ablation_attenton}
    \scriptsize \centering
    \begin{tabular}{lccccccc}
    \toprule
      Method  & SC ($\uparrow$) & BC ($\uparrow$) & MS ($\uparrow$) & IQ ($\uparrow$) & AQ ($\uparrow$)& DD ($\uparrow$)\\
    \midrule
    Sliding Window & 97.52& 96.72 & 98.69 & 64.33& 54.92& \textbf{37.81}\\
    Selected Frame Attention & 97.89 & 98.94 & 98.94& 63.90& 54.19 & 28.53 \\
    Our TSA& \cellcolor{mitblue}\textbf{98.44
    }& \cellcolor{mitblue}\textbf{97.78} & \cellcolor{mitblue}\textbf{98.97} & \cellcolor{mitblue}\textbf{68.84} & \cellcolor{mitblue}\textbf{61.21} & \cellcolor{mitblue}36.27 \\
    \bottomrule
    \end{tabular}
    \vspace{-15pt}
\end{table}

%% file: sec/5_conclusion.tex
\section{Conclusion}
\label{sec:conclusion}
In this paper, we introduced FreeLOC, a layer-adaptive, training-free framework that enables pre-trained video DiTs to generate high-fidelity long videos. Our approach systematically resolves two  O.O.D issues by using VRPR to remap temporal positions and TSA to preserve attentional focus. The core of FreeLOC is a layer-adaptive strategy, guided by an automatic probing mechanism, which applies these techniques selectively for optimal performance. Extensive experiments demonstrate that FreeLOC significantly surpasses existing training-free approaches, offering a powerful and generalizable solution for long video synthesis.

%% file: sec/X_suppl.tex
\clearpage
\setcounter{page}{1}
\setcounter{figure}{0}
\setcounter{table}{0}
\setcounter{equation}{0}
\setcounter{section}{0}
\maketitlesupplementary
\setcounter{tocdepth}{2} %
\tableofcontents         %
\vspace{0.5cm}
\addtocontents{toc}{\protect\setcounter{tocdepth}{2}}

\section{Verification of Context-Length O.O.D}

As mentioned in the main paper, we find that naively extending the frame-level context length (with VRPR applied to constrain the frame-level relative position within pretrained range) leads to increased attention entropy in the video DiT. This diffusion of attention weights correlates with a degradation in generation quality.

To quantify this, we measure the attention entropy. Assume that we generate a video sequence with a total token number of $N$, where $N = f \times n$ ($f$ is the total number of frames and $n$ is the number of tokens per frame). Let $p_1, p_2, \cdots, p_N$ be the sequence of attention distributions (after softmax) for all tokens. Each $p_i$ is a vector of attention weights for the $i$-th token, and this vector has a length of $N$.
We compute the standard Shannon entropy for the attention distribution of each token $i$ as:
\begin{equation}
    H(p_i) = - \sum_{j=1}^{L} p_{i,j} \ln(p_{i,j})
\small
\end{equation}
To obtain the final average attention entropy, we first compute the mean of $H(p_i)$ over all $L$ tokens, and then average this value across all attention heads and all self-attention layers.

Figure~\ref{fig:contex_length_ood} plots the resulting average attention entropy as it varies with the context length (i.e., the total number of frames, $f$). This illustrates that as the context length increases, the attention entropy also rises, confirming the context-length O.O.D phenomenon.

\input{figure_tex/context_length_ood}

\section{Additional Details of VRPR and TSA}

\subsection{Implementation Details of VRPR}
As stated in the main paper, VRPR is designed to remap O.O.D relative positions back into the pre-trained distribution in a multi-granularity fashion.

The re-encoding functions (Equations 2 and 3 in the main paper) are designed to create a "compressed" representation of relative time.
Specifically, for a model pre-trained on $L$ frames (e.g., $L=81$ for Wan2.1-T2V-1.3B), the pre-trained relative position range is $[-(L-1), L-1]$. 

Here, we provide the specific implementation details for Video-based Relative Position Re-encoding (VRPR). Directly modifying the relative position matrix as described in the main paper is practically challenging,
To address this, we propose an approximate implementation achieved through the formulation of region-specific position indices. Let $P_q[i]$ denote the position index for the query at frame $i$ after re-encoding,  $P_k[j]$ denote the position index for the key at frame $j$ after re-encoding and $P_\text{impl} $ denote the implemented relative position.

\noindent \textbf{1. Fine-Grained Re-Encoding}:\\
For local interactions within the fine-grained window  ($|i-j| \le W_1$), we use standard relative positions:
\begin{equation}
    P_q[i] = i, \quad P_k[j] = j
    \small
\end{equation}
In this case, the effective relative position is $P_\text{impl} = P_q[i] - P_k[j] = i - j = P$. This provides an exact implementation of the identity mapping $P = P_\text{ori}$ described in the main paper.

\noindent \textbf{2. Medium-Grained Re-Encoding}:\\
For interactions in the medium range ($W_1 < i-j \le W_2$), we apply the following mapping:
\begin{equation}
    P_q[i] = \left\lfloor \frac{i}{G_1} \right\rfloor + \left(W_1 - \left\lfloor\frac{W_1}{G_1}\right\rfloor\right), \quad P_k[j] = \left\lfloor \frac{j}{G_1} \right\rfloor
    \small
\end{equation}
Conversely, for negative medium-range distances ($-W_2 \le i-j < -W_1$):
\begin{equation}
    P_q[i] = \left\lfloor \frac{i}{G_1} \right\rfloor, \quad P_k[j] = \left\lfloor \frac{j}{G_1} \right\rfloor + \left(W_1 - \left\lfloor\frac{W_1}{G_1}\right\rfloor\right)
    \small
\end{equation}
This formulation approximates the theoretical formula presented in the main paper (Eq. 2). Specifically, for the positive case, the relative position calculated by this implementation is:
\begin{equation}
    P_\text{impl} = P_q[i] - P_k[j] = \left(\left\lfloor \frac{i}{G_1} \right\rfloor - \left\lfloor \frac{j}{G_1} \right\rfloor\right) + \left(W_1 - \left\lfloor\frac{W_1}{G_1}\right\rfloor\right)
    \small
\end{equation}
Comparing this to the main paper's target formula $P = \lfloor \frac{i-j}{G_1} \rfloor + (W_1 - \lfloor\frac{W_1}{G_1}\rfloor)$, the equivalence relies on the mathematical property that
\begin{equation}
  \left\lfloor \frac{i}{G_1} \right\rfloor - \left\lfloor \frac{j}{G_1} \right\rfloor - 1 \le  \left\lfloor \frac{i-j}{G_1} \right\rfloor \le  \left\lfloor \frac{i}{G_1} \right\rfloor -  \left\lfloor \frac{j}{G_1} \right\rfloor.
    \small
\end{equation}
So, the error between $P_\text{impl}$ and $P$ is $ P_\text{impl}-P=\left\lfloor \frac{i-j}{G_1} \right\rfloor-(\left\lfloor \frac{i}{G_1} \right\rfloor -  \left\lfloor \frac{j}{G_1} \right\rfloor) \in \{0,-1\}$.
This approximation error is negligible for our case and allows for the decomposition of relative positions into independent query and key indices. Crucially, this approximation preserves the original granularity (i.e., the group size)  and the monotonicity of relative position while effectively constraining the relative position within the pre-trained length, preventing O.O.D issues. With this operation, it also guarantees no discontinuity near the boundary of different regions.

\noindent \textbf{3. Coarse-Grained Re-Encoding}:

For distant interactions ($i-j > W_2$), we employ a coarser quantization:
\begin{equation}
    P_q[i] = \left\lfloor \frac{i}{G_2} \right\rfloor + \left(W_2 - \left\lfloor \frac{W_2}{G_2} \right\rfloor - \left\lfloor \frac{W_2-W_1}{G_1} \right\rfloor\right), P_k[j] = \left\lfloor \frac{j}{G_2} \right\rfloor
    \small
\end{equation}
For the negative direction ($i-j < -W_2$):
\begin{equation}
     P_q[i] = \left\lfloor \frac{i}{G_2} \right\rfloor, P_k[j] = \left\lfloor \frac{j}{G_2} \right\rfloor + \left(W_2 - \left\lfloor \frac{W_2}{G_2} \right\rfloor - \left\lfloor \frac{W_2-W_1}{G_1} \right\rfloor\right)
    \small
\end{equation}

Importantly, our dynamic mapping strategy preserves monotonicity property of relative position across region boundaries: tokens at greater distances consistently receive larger absolute relative position values. Furthermore, this index-based formulation allows VRPR to seamlessly integrate with efficient attention implementations such as FlashAttention2 following~\cite{jin2024llm}, as it only requires modifying the input position indices rather than the full attention matrix.

The specific hyperparameter configurations for the models evaluated in the main paper are as follows. These parameters include the Local Window size ($W_1$), the Mid-Range Window size ($W_2$), Group Size 1 ($G_1$) and Group Size 2 ($G_2$).

\begin{figure*}[t]
  \centering
  \vspace{-10pt}
  \includegraphics[width=1\linewidth]{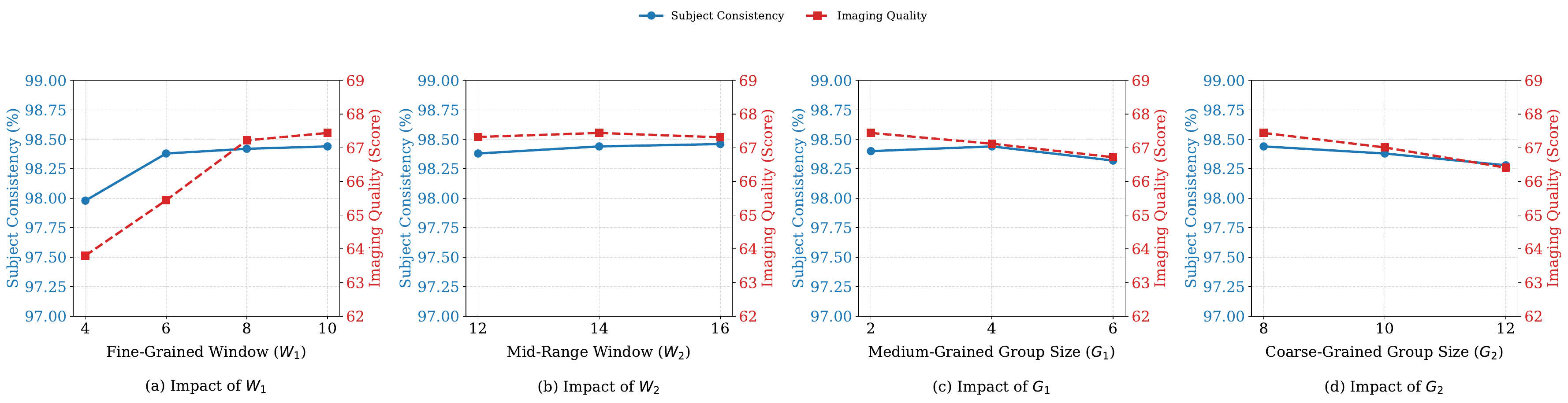}
   \caption{\textbf{Impact of $W_1$, $W_2$ , $G_1$ and $G_2$ on Subject Consistency and Imaging Quality for VRPR}.}
   \label{fig:vrpr}
   \vspace{-10pt}
\end{figure*}

The principles of hyperparameters selection is that the maximum relative position in the generated video must not exceed the pre-trained context window $L$. For a target generation length $L_\text{target}$, the maximum absolute relative distance is $ L_\text{target} - 1$. The mapped position $P(L_\text{target} - 1)$ must satisfy:
\begin{equation}
    P_{\text{impl}}(L_\text{target} - 1) \le L - 1.
    \small
\end{equation}
Substituting the coarse-grained mapping (assuming $D_{max} > W_2$), the criterion is formalized as:
\begin{equation}
    \left\lfloor \frac{L_\text{target} - 1}{G_2} \right\rfloor + \left(W_2 - \left\lfloor \frac{W_2}{G_2} \right\rfloor - \left\lfloor \frac{W_2-W_1}{G_1} \right\rfloor\right) \le  L- 1.
    \small
    \label{eq:inequal}
\end{equation}
We select the possible windows $W_1, W_2$ and group sizes $G_1, G_2$ that satisfy this inequality to  prevent O.O.D issues. The empirical configurations for our experiments are:

\noindent\textbf{For Wan2.1-T2V-1.3B:}
\begin{itemize}[leftmargin=*]
    \item \textbf{$2\times$Extension (161-frame):} Local Window $W_1 = 12$, Mid-Range Window $W_2 = 20$, Group Size 1 $G_1=2$ and Group Size 2 $G_2=8$
    \item \textbf{$4\times$Extension (321-frame):} Local Window $W_1 = 10$, Mid-Range Window $W_2 = 14$, Group Size 1 $G_1=2$ and Group Size 2 $G_2=8$.
\end{itemize}
\noindent\textbf{For HunyuanVideo:}
\begin{itemize}[leftmargin=*]
    \item \textbf{$2\times$Extension (253-frame):} Local Window $W_1 = 12$, Mid-Range Window $W_2 = 20$, Group Size 1 $G_1=2$ and Group Size 2 $G_2=4$. 
    \item \textbf{$4\times$Extension (509-frame):} Local Window $W_1 = 12$, Mid-Range Window $W_2 = 20$, Group Size 1 $G_1=2$ and Group Size 2 $G_2=8$.
\end{itemize}
These values were determined empirically to provide a good balance between preserving local motion and maintaining global coherence. Noting that we also observed that as long as these parameters satisfy the aforementioned inequality~\ref{eq:inequal}, slight adjustments to them (modify a single parameter at a time while maintaining the other three fixed) have minimal impact on the results (performance metrics) as shown on Figure~\ref{fig:vrpr}, demonstrating relatively good robustness.

\input{figure_tex/attention_mask}

\subsection{Implementation Details of TSA}
TSA is designed to combat the context-length O.O.D problem by preserving attention density. It structures the attention mask into three components. Here, we provide an enlarged and more detailed view of the attention mask in Figure~\ref{fig:attn_mask}. 

\textbf{Detailed Derivation of Striped Window Width ($D_S=\lfloor\frac{nD_1}{\alpha(D_2-D_1)}\rfloor$)}.
A key innovation of TSA is the adaptive calculation of the spatial window width $D_s$ ( which is not explored in detail in the main text). Our theoretical goal is to strictly control the "Attention Mass" (the total number of key tokens involved in calculation) in the mid-range zone to prevent it from dominating the local attention.
Let $N_\text{local}$ be the total number of tokens attended to in the dense local window. Given $n$ tokens per frame and a local window size of $D_1$, the local token number (attention density) is:
\begin{equation}
    N_\text{local} = n \times D_1.
\end{equation}
For the mid-range zone spanning $(D_2 - D_1)$ frames, using dense attention would result in a token count of $n \times (D_2 - D_1)$, which is typically too large. We introduce the $\alpha$, which defines the decay ratio. Specifically, we mandate that the total attention density in the mid-range ($N_\text{mid}$) should be scaled down by $\alpha$ relative to the local density:
\begin{equation}
    N_\text{mid} = \frac{N_\text{local}}{\alpha}.
\end{equation}
Given the spatial strip width $D_s$, the actual token count in the mid-range is $N_\text{mid} = D_s \times (D_2 - D_1)$. Equating these terms allows us to derive the optimal $D_s$:
\begin{equation}
    D_s \times (D_2 - D_1) = \frac{n \times D_1}{\alpha} \implies D_s = \left\lfloor \frac{n D_1}{\alpha (D_2 - D_1)} \right\rfloor
    \small.
\end{equation}
The parameter $\alpha$ explicitly controls the trade-off between global context awareness and local detail preservation. $\alpha$ represents the ratio of "Local Attention Density" to "Mid-Range Attention Density". For example, $\alpha=2$ implies that the aggregate information retrieved from the extended temporal context ($D_1$ to $D_2$) is compressed to exactly half the capacity of the immediate local context. By setting $\alpha > 1$, we force the model to prioritize local interactions (which define motion quality) while treating mid-range interactions as supplementary cues.

The specific hyperparameter configurations for the models evaluated in the main paper are as follows. These parameters include the Local Window size ($D_1$), the Mid-Range Window size ($D_2$), and the Striped Attention Density ($\alpha$). 

In principle, firstly, the selection of $D_1$ and $\alpha$ should adhere to the following requirements: $nD_1 +  \lfloor\frac{nD_1}{\alpha(D_2-D_1)}\rfloor(D_2-D_1) \le nD_1+\lfloor\frac{nD_1}{\alpha}\rfloor\le nD_1(1+\frac{1}{\alpha})\le \frac{nL_\text{pretrained}}{2}$, i.e., the effective token number stays within the pretrained range, where $L_\text{pretrained}$ is the pretrained frame length. Consequently, $D_1$ and $\alpha$ only need to satisfy  $D_1(1+\frac{1}{\alpha})\le \frac{L_\text{pretrained}}{2}$. However, an excessively small $D_1$ would result in an overly limited interaction range for the attention mechanism, which is undesirable for temporal consistency. So we empirically impose the additional constraint that: 
\begin{equation}
\label{eq:tsa}
    \frac{L_\text{pretrained}}{4} \le D_1(1+\frac{1}{\alpha})\le \frac{L_\text{pretrained}}{2}.
\end{equation}
For Wan2.1-T2V-1.3B, we set $L_\text{pretrained}=24$, and for HunyuanVideo, we set $L_\text{pretrained}=36$. For the selection of $D_2$, it only has to satisfy $D_2>D_1$, and we will discuss effect of its choice in following ablation study.

Specifically,

\noindent\textbf{For Wan2.1-T2V-1.3B:}
\begin{itemize}[leftmargin=*]
    \item \textbf{$2\times$Extension (161-frame):} Local Window $D_1 = 8$, Mid-Range Window $D_2 = 16$, and Striped Attention Density $\alpha = 4$.
    \item \textbf{$4\times$Extension (321-frame):} Local Window $D_1 = 8$, Mid-Range Window $D_2 = 24$, and Striped Attention Density $\alpha = 4$.
\end{itemize}
\noindent\textbf{For HunyuanVideo:}
\begin{itemize}[leftmargin=*]
    \item \textbf{$2\times$Extension (253-frame):} Local Window $D_1 = 12$, Mid-Range Window $D_2 = 24$, and Striped Attention Density $\alpha = 4$.
    \item \textbf{$4\times$Extension (509-frame):} Local Window $D_1 = 12$, Mid-Range Window $D_2 = 36$, and Striped Attention Density $\alpha = 4$.
\end{itemize}
The \textbf{Attention Sink} component is implemented by ensuring that all tokens (queries) from all frames are permitted to attend to the tokens from the initial frame.

\section{Layer-wise Sensitivity Profile and Strategy}
To rigorously determine the layer-wise application of FreeLOC, we conducted a comprehensive probing experiment. The layer-wise probing was conducted using $N=10$ diverse prompts, generating $M=3$ videos per prompt for each perturbation configuration to ensure statistical reliability.

\textbf{Sensitivity Determination.} 
We identify layers sensitive to frame-level relative position O.O.D based on VisionReward and Attention Logits Difference (ALD), and layers sensitive to context-length O.O.D based on the Context-Length Sensitivity Score. To establish a definitive binary sensitivity classification, we employ a thresholding strategy where the top two-thirds  of layers exhibiting the most significant metric degradation (or score increase) are designated as sensitive to each respective O.O.D type.

\noindent\textbf{Profile for Wan2.1-T2V-1.3B.}
Based on this protocol, the specific sensitivity profile for the 30 layers of Wan2.1-T2V-1.3B is identified as follows:
\begin{itemize}
    \item \textbf{Layers sensitive to Frame-level Relative Position O.O.D (20 layers):} \\ $\{28,1,29,0,27,26,21,17,20,12,16,14,8,4,13,19,10,\\23,5,3\}$.
    \item \textbf{Layers sensitive to Context-Length O.O.D (20 layers):} \\ $\{18, 24, 13, 23, 22, 10, 15, 6, 11,25,2,4,16,29,7, 8, 12\\9,3,21\}$.
\end{itemize}
The sequence here is arranged in descending order of sensitivity.
Notably, we conclude that in this configuration, \textit{every single layer} is sensitive to at least one type of O.O.D issue, necessitating a comprehensive correction strategy.

\textbf{Layer-wise Strategy}
Given these overlapping sensitivities, we devised a balanced allocation strategy to optimize performance.
\begin{enumerate}
    \item For layers identified as sensitive to \textbf{Frame-level Relative Position O.O.D}, we apply the \textbf{VRPR} strategy.
    \item For layers identified as sensitive to \textbf{Context-Length O.O.D}, we apply the combined \textbf{VRPR + TSA} strategy.
\end{enumerate}
We found that an even distribution—allocating the top 50\% of layers (specifically, the 15 layers most sensitive to context length) to the VRPR+TSA strategy, and the remaining 50\% to the VRPR-only strategy—yields the relatively optimal balance. Interestingly, the 16 layers identified as context-sensitive in our probing aligns almost perfectly with this 50\% allocation.

\begin{figure*}[h]
  \centering
  \includegraphics[width=1\linewidth]{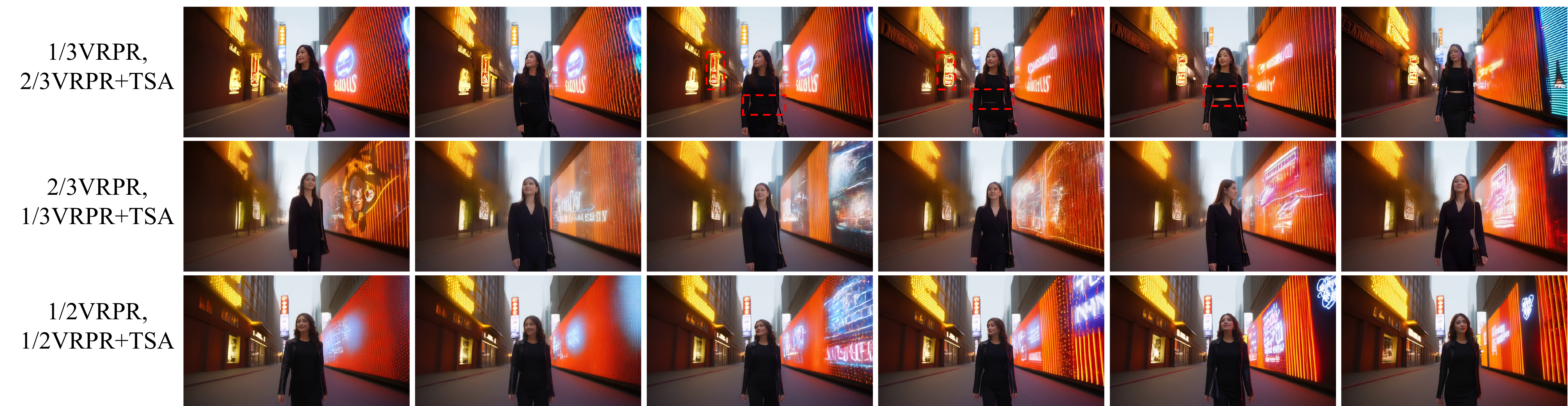}
   \caption{\textbf{Qualitative comparison of different allocation strategy}. The (1/2, 1/2) split proved to be the robust "sweet spot" which balance detail preservation and temporal consistency.}
   \label{fig:proportion}
\end{figure*}

To validate this choice, we compared this balanced (1/2 VRPR, 1/2 VRPR+TSA) distribution against other ratios:
\begin{itemize}
    \item \textbf{1/3 VRPR, 2/3 VRPR+TSA}: Resulted in over-sparsification, harming local details.
    \item \textbf{2/3 VRPR, 1/3 VRPR+TSA}: Insufficient context correction, leading to visual details degradation in long videos.
\end{itemize}
The 50\%/50\% split proved to be the robust "sweet spot". The detailed visualization of Qualitative comparison is shown in Figure~\ref{fig:proportion}.

\noindent\textbf{Profile for Hunyuan Video.}
Resembling the probing procdedure of Wan2.1-T2V-1.3B, the specific sensitivity profile for the 60 layers of Hunyuan Video is identified as follows:

\begin{itemize}
    \item \textbf{Context-length sensitive layers (40 layers):} These layers exhibit high sensitivity to context length and first 30 layers are assigned the \textbf{VRPR+TSA} strategy. The identified layers are: \\
    $\{1, 2, 45, 29, 5, 13, 38, 59, 11, 34, 47, 16, 0, 53, 21, 41, \\ 8, 30, 55, 19, 4, 49, 26, 42, 12, 57, 23, 25, 36, 51, 15, 39, 7, \\22, 44, 18, 32, 9, 27, 35\}$.
    
    \item \textbf{Frame-level Relative Position-sensitive layers (40 layers):} These layers exhibit sensitivity to relative position shifts. Layers that appear in this list \textit{but not} in the context-length sensitive list are assigned the \textbf{VRPR only} strategy. The identified layers are: \\
    $\{58, 59, 31, 6, 54, 21, 37, 29, 47, 18, 28, 10,  1, 24, 3,
    \\35, 41, 14, 52, 27, 9, 43, 17, 33, 48, 
    2, 23, 50, 11, 38, 4, 25, \\40, 15, 0, 7, 56, 20, 46, 32\}$.
\end{itemize}

\section{More Implementation Details}

All experiments were conducted on a single NVIDIA A100 (80GB) GPU. Basic settings for generation are listed in Table \ref{tab:hyperparams}. For test prompt, we randomly sample 100 prompts from Vbench-long~\cite{huang2024vbench} following prior works~\cite{lu2024freelong, qiu2023freenoise, zhao2025riflex}.

\begin{table}[t!]
\centering
\caption{\textbf{Generation Settings.} Basic settings used for Wan2.1-T2V-1.3B and Hunyuan Video for quantitative and qualitative evaluation.}
\label{tab:hyperparams}
\begin{tabular}{@{}lll@{}}
\toprule
Parameter & Wan2.1-T2V-1.3B & Hunyuan Video \\ \midrule
Sampler & Unipc& Euler \\
Denoising Steps & 50 & 50 \\
CFG Scale & 6.0 & 6.0 \\
Resolution & $832 \times 480$ (480p) & $832 \times 480$ (480p) \\ \midrule
Base Length (1x) & 81 frames & 127 frames \\
2x Length & 161 frames & 253 frames \\
4x Length & 321 frames & 509 frames \\ \bottomrule
\end{tabular}
\end{table}

\section{Additional Ablation Study Results}
For efficiency, we conduct following ablation studies on Wan2.1-T2V-1.3B~\cite{wan2025wan} with $4\times$ length extension (similar results can also be obtained on HunyuanVideo).

\subsection{Impact of $D_1$, $D_2$ and $\alpha$ for TSA}
To investigate the individual contributions of the TSA hyperparameters—local window size $D_1$, mid-range window boundary $D_2$, and attention decay factor $\alpha$—we conducted controlled ablation studies measuring Subject Consistency (SC) to measure temporal consistency and Imaging Quality (IQ) to measure video quality. We conduct following experiments: (1) \textbf{Impact of $D_1$:} Fixing $D_2=24$ and $\alpha=4$, we varied $D_1 \in [5, 9]$ according to Eq~\ref{eq:tsa}. (2) \textbf{Impact of $D_2$:} Fixing $D_1=8$ and $\alpha=4$, we varied $D_2 \in [8, 16, 24, 32, 40, 68]$. (3) \textbf{Impact of $\alpha$:} Fixing $D_1=8$ and $D_2=24$, we varied $\alpha \in [1, 2, 4, 6, 8]$.

\begin{figure*}[t]
  \centering
  \vspace{-10pt}
  \includegraphics[width=1\linewidth]{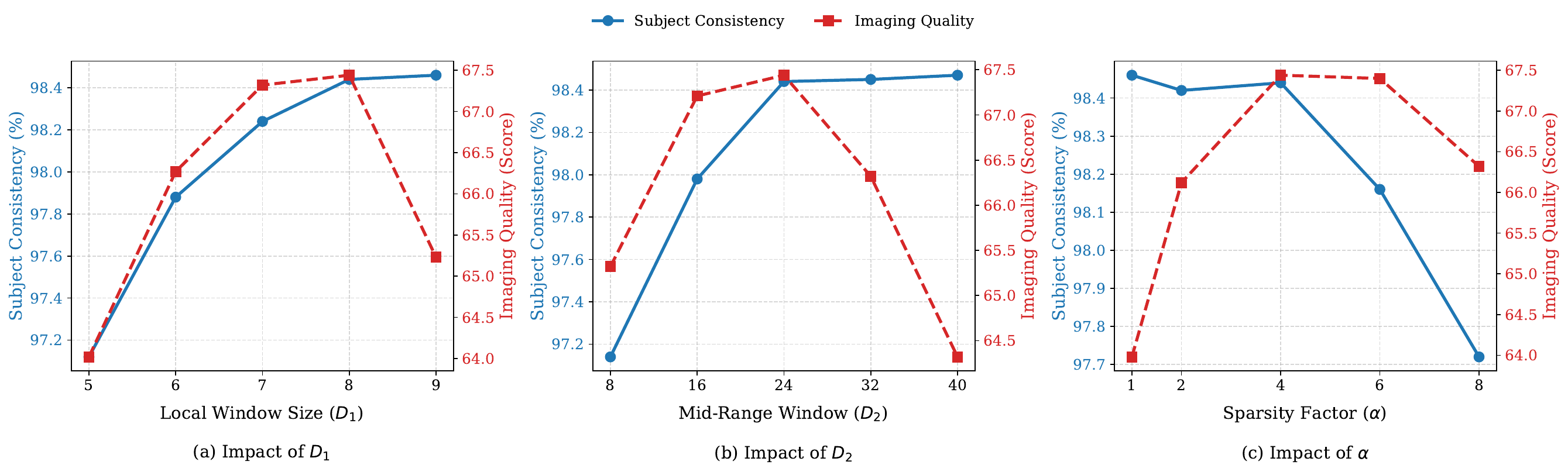}
   \caption{\textbf{Impact of $D_1$, $D_2$ and $\alpha$ on Subject Consistency and Imaging Quality}.}
   \label{fig:tsa_ablation}
   \vspace{-10pt}
\end{figure*}

\textbf{Results Analysis:}
Figure \ref{fig:tsa_ablation} illustrates the trade-offs observed:
\begin{itemize}
    \item \textbf{Varying $D_1$:} As $D_1$ increases, SC improves continuously due to stronger local temporal coupling. However, IQ initially improves but subsequently declines, likely because an overly large dense window introduces excessive attention entropy.
    \item \textbf{Varying $D_2$:} Increasing $D_2$ initially boosts SC, which then plateaus as the benefit of mid-range context saturates. IQ follows an inverted-U pattern, increasing at first but dropping if the mid-range window becomes too computationally diffuse.
    \item \textbf{Varying $\alpha$:} As $\alpha$ increases (sparser attention), SC remains stable initially but eventually drops as temporal connections become too sparse. Conversely, IQ improves continuously and then plateaus, as sparsity effectively reduces noise and attention blurring.
\end{itemize}

\subsection{Impact of Attention Sink}

Attention sink is not unique to LLMs and prior analyses of video DiTs' attention maps also show sink behavior that serves as a global anchor to prevent identity drift instead of introducing a static bias. We run simple ablations with mid-frame sink, last-frame sink and without attention sink in Table~\ref{tab:attn_sink}, and first-frame sink performs best. For videos with large scene changes, such sink may be suboptimal, but our method can extend to adaptive sink to support this case and we will improve it in the future.

\begin{table}[t]
\centering
    \setlength{\tabcolsep}{4pt}
    \caption{\textbf{Ablation study of impact of attention sink}. Attention could significantly improve overall video consistency and video quality.}
    \label{tab:attn_sink}
    \scriptsize \centering
    \begin{tabular}{lccccccc}
    \toprule
      Method  & SC ($\uparrow$) & BC ($\uparrow$) & MS ($\uparrow$) & IQ ($\uparrow$) & AQ ($\uparrow$)& DD ($\uparrow$)\\
    \midrule
    Mid Frame & 98.14 & 97.65 &98.91 & 67.12 & 59.83 & 34.65 \\
    Last Frame & 98.19 & 97.68 &98.87 & 67.48 & 60.98 & 35.72 \\
    w/o Attention Sink & 98.05 &97.67 & 98.92 & \textbf{67.53} & 60.92& 34.59 \\
    \cellcolor{mitblue}\textbf{Our First Frame}& \cellcolor{mitblue}\textbf{98.44
    }& \cellcolor{mitblue}\textbf{97.78} & \cellcolor{mitblue}\textbf{98.97} & \cellcolor{mitblue}67.44 & \cellcolor{mitblue}\textbf{61.21} & \cellcolor{mitblue}\textbf{36.27} \\
    \bottomrule
    \end{tabular}
 \end{table}

\subsection{More Comparison with Other RoPE Scaling Methods}
We compare VRPR with other RoPE scaling techniques e.g., LI (Linear Interpolation), NTK-aware scaling and YaRN in Table~\ref{tab:rope}. 
Results confirm that standard RoPE scaling methods degrade visual details by diluting local positional frequencies, proving VRPR's hierarchical preservation of local precision is essential for high-quality generation.

\begin{table}[t]
\centering
    \setlength{\tabcolsep}{6pt}
    \caption{\textbf{VRPR vs. other RoPE scaling methods.}}
    \label{tab:rope}
    \scriptsize \centering
    \begin{tabular}{lccccccc}
    \toprule
      Method  & SC ($\uparrow$) & BC ($\uparrow$) & MS ($\uparrow$) & IQ ($\uparrow$) & AQ ($\uparrow$)& DD ($\uparrow$)\\
    \midrule
    LI & 97.93 & 97.62 &98.64 & 51.11 & 50.91 & 16.32 \\
    YARN & 98.16 & 97.64 & 98.89 & 62.33  & 57.98 & 30.87 \\
    NTK-aware & 97.89 & 97.59 & 98.82 & 56.47 & 53.54 & 24.21\\
    Our VRPR& \cellcolor{mitblue}\textbf{98.44
    }& \cellcolor{mitblue}\textbf{97.78} & \cellcolor{mitblue}\textbf{98.97} & \cellcolor{mitblue}\textbf{68.84} & \cellcolor{mitblue}\textbf{61.21} & \cellcolor{mitblue}\textbf{36.27} \\
    \bottomrule
    \end{tabular}
    \scriptsize
 \end{table}

\subsection{Different Granularity of VRPR}
We validated the necessity of the 3-stage VRPR design (Fine $\rightarrow$ Medium $\rightarrow$ Coarse) against a 2-stage baseline which (Fine $\rightarrow$ Coarse) and a 4-stage baseline (Fine $\rightarrow$ Medium 1 $\rightarrow$ Medium 2 $\rightarrow$ Coarse). All these designs re-encode frame-level relative position within pretrained range. 
The result is shown in Table~\ref{tab:different_granularity}. The obtained results demonstrate that employing a multi-granularity (three granularities) approach, as opposed to a strategy with only two granularities, leads to improved video consistency and quality. However, diminishing returns are observed when the granularity continues to increase (e.g., four levels of granularity).

\begin{table}[h]
\centering
    \vspace{-5pt}
    \setlength{\tabcolsep}{5.5pt}
    \caption{\textbf{Ablation study of different granularity of VRPR.}}
    \vspace{-5pt}
    \label{tab:different_granularity}
    \scriptsize \centering
    \begin{tabular}{lccccccc}
    \toprule
      Method  & SC ($\uparrow$) & BC ($\uparrow$) & MS ($\uparrow$) & IQ ($\uparrow$) & AQ ($\uparrow$)& DD ($\uparrow$)\\
    \midrule
    2-stage & 98.15 &97.69 & 98.90 & 65.33 & 58.91& 33.42 \\
    4-stage & \textbf{98.49} &97.93 & 98.95 & \textbf{67.91} & 60.90& 35.97 \\
    \cellcolor{mitblue}\textbf{Our 3-stage}& \cellcolor{mitblue}\textbf98.44
    & \cellcolor{mitblue}\textbf{97.78} & \cellcolor{mitblue}\textbf{98.97} & \cellcolor{mitblue}67.44 & \cellcolor{mitblue}\textbf{61.21} & \cellcolor{mitblue}\textbf{36.27} \\
    \bottomrule
    \end{tabular}
    \vspace{-5pt}
 \end{table}

\subsection{More Layer-wise Strategy}
To further verify the correctness and effectiveness of our layer-wise adaptive strategy, we conducted more experiments on different layer-wise strategy. (1) "Reverse" layer-wise strategy, we inverted the assignment logic: applying TSA only to layers identified as \textit{not} sensitive to context length, and applying VRPR+TSA only to layers \textit{not} sensitive to position. (2) Half Implementation 1: applying VRPR+TSA to the first half of the layers and VRPR to the second half.(3) Half Part Implementation 2: applying VRPR to the first half of the layers and VRPR+TSA to the second half. (4) Interleaved: alternately applying VRPR and VRPR+TSA strategies layer by layer.

As shown in Table \ref{tab:more_layer_wise}, the Reverse configuration, Half Implementation and Interleaved configuration lead to a significant drop in both Subject Consistency (SC) and Imaging Quality (IQ), validating that our probing mechanism accurately identifies the specific needs of each layer.

\begin{table}[h]
\centering
    \vspace{-5pt}
    \setlength{\tabcolsep}{4pt}
    \caption{\textbf{Ablation study of more different layer-wise strategy.}}
    \vspace{-5pt}
    \label{tab:more_layer_wise}
    \scriptsize \centering
    \begin{tabular}{lccccccc}
    \toprule
      Method  & SC ($\uparrow$) & BC ($\uparrow$) & MS ($\uparrow$) & IQ ($\uparrow$) & AQ ($\uparrow$)& DD ($\uparrow$)\\
    \midrule
    Reverse & 98.02 &97.28 & 98.87 &62.12 & 57.12& 31.23 \\
     Half Implementation 1 & 98.18 &97.41 & 98.91 & 64.21 & 59.12& 32.19 \\
     Half Implementation 2 & 98.22 &97.48 & 98.94 & 62.39 & 60.91& 32.92 \\
     Interleaved & 98.12 &97.32 & 98.89 &  63.20 & 59.92& 33.59 \\
    \cellcolor{mitblue}\textbf{FreeLOC}& \cellcolor{mitblue}\textbf{98.44
    }& \cellcolor{mitblue}\textbf{97.78} & \cellcolor{mitblue}\textbf{98.97} & \cellcolor{mitblue}\textbf{67.44} & \cellcolor{mitblue}\textbf{61.21} & \cellcolor{mitblue}\textbf{36.27} \\
    \bottomrule
    \end{tabular}
    \vspace{-5pt}
 \end{table}

\section{More Experiment Results}

\subsection{Inference Efficiency Analysis}

In this section, we provide a comprehensive comparison of efficiency, focusing on both inference time (measured as the time required for each denoising step) and peak GPU memory usage. We compare our method against other training-free methods, along with the \textit{Direct Sampling} and \textit{Sliding Window} baselines.
For this comparison, all methods were applied to the Wan2.1-T2V-1.3B~\cite{wan2025wan} model to generate 321-frame videos. All measurements were conducted on an NVIDIA A100 GPU. As presented in Table~\ref{tab:inference_time}, while dramatically improving the quality of long videos generated by short video models, our method introduces no significant increase in inference time or peak memory consumption. This efficiency is primarily attributed to the optimized computations resulting from the Tiered Sparse Attention (TSA) mechanism, which effectively manages the computational cost and memory footprint associated with long-context attention.

\begin{table}[htbp]
\vspace{-5pt}
  \centering
  \caption{\textbf{Comparison of inference time and peak GPU memory usage} for generating 321-frame videos on the Wan2.1-T2V-1.3B model. All measurements were conducted on an NVIDIA A100 GPU.}
  \label{tab:inference_time}
  \vspace{-5pt}
  \begin{tabular}{lcc}
    \toprule
    Method & Inference Time (s/step) & Mem (GB) \\
    \midrule
    Direct Sampling & 33.93 &  29.34 \\
    Sliding Window & 16.72 & 30.33\\
    FreeNoise~\cite{qiu2023freenoise} & 17.21 &30.34\\
    Freelong~\cite{lu2024freelong} & 48.15 & 40.71\\
    RIFLEX~\cite{zhao2025riflex} & 33.94 & 29.34\\
    \cellcolor{mitblue}\textbf{FreeLOC} &  \cellcolor{mitblue} 24.35&  \cellcolor{mitblue} 29.87\\
    \bottomrule
  \end{tabular}
\end{table}

\subsection{One-Time Cost and Negligible Compute}
Our probing is a one-time architectural characterization: the layer-wise sensitivity profile is consistent across random seeds, prompts and aspect ratios, and is fixed for that model once obtained.
Although the main paper reports generating $M\times N$ videos for completeness,
we empirically find that the profile is highly stable and does not require the full probing set.
Specifically on Wan2.1-T2V-1.3B~\cite{wan2025wan}, using a single prompt to derive the profile yields high Spearman rank correlation (evaluated over 10 randomly sampled single prompts) of ($\rho = 0.984 \pm 0.009$) compared to the full profile.
Thus, a single sample set is therefore sufficient in practice, rendering the probing cost negligible. For Wan2.1-T2V-1.3B, probing requires only 4 hours on an RTX 4090. Moreover, \emph{we will release pre-computed sensitivity profiles for popular models}, eliminating the need for users \emph{to rerun probing}.

\begin{table}[!t]
  \centering
  \small
  \caption{\textbf{User study results}. Participants scored videos on a scale of 1 (worst) to 5 (best) across three criteria. Our method received the highest ratings across all three metrics.}
  
  \label{tab:user_study}
  \setlength{\tabcolsep}{3pt}
  \begin{tabular}{lccc}
    \toprule
    Method & \begin{tabular}[c]{@{}c@{}}Content \\ Consistency\end{tabular} & \begin{tabular}[c]{@{}c@{}}Video \\ Quality\end{tabular} & \begin{tabular}[c]{@{}c@{}}Video-Text \\ Alignment\end{tabular} \\
    \midrule
    Direct Sampling &  3.27 & 1.02 & 1.92 \\
    Sliding Window & 1.17 & 3.25 &  2.34\\
    FreeNoise~\cite{qiu2023freenoise} & 1.20 &2.88  & 2.68 \\
    Freelong~\cite{lu2024freelong}  & 3.59 & 2.11 & 2.98\\
    RIFLEX~\cite{zhao2025riflex} & 3.31 &1.09 & 1.89 \\
    \cellcolor{mitblue}\textbf{FreeLOC} &\cellcolor{mitblue} \textbf{4.11} &\cellcolor{mitblue} \textbf{3.95}  & \cellcolor{mitblue} \textbf{3.78} \\
    \bottomrule
  \end{tabular}
\end{table}
\subsection{User Study}

To further assess our results based on human subjective judgment, we carried out a user study. In this study, participants were shown long videos generated using Wan2.1-T2V-1.3B as the base short video model. Videos from all compared methods were included, totaling 50 videos. The examples were presented to participants in a random order to eliminate potential bias.

Participants were asked to score the generated videos on a scale of 1 to 5 according to three evaluation criteria: (1) \textbf{Content Consistency}, (2) \textbf{Video Quality}, and (3) \textbf{Video-Text Alignment}. The average scores for each method are reported in Table~\ref{tab:user_study}. As shown, our method received the highest ratings across all three metrics.

\section{More Qualitative Results}
We provide extensive qualitative comparisons to visually demonstrate the superiority of FreeLOC. Figures~\ref{fig:quality_wan_321},~\ref{fig:quality_wan_161},~\ref{fig:quality_hunyuan_321} and~\ref{fig:quality_hunyuan_161} show additional qualitative results on Wan2.1-T2V-1.3B and Hunyuan Video. Our method consistently produces videos with higher temporal fidelity, sharper details, and more stable object identity compared to all baselines.

\input{figure_tex/figure_quality_sup/quality_wan_321}
\input{figure_tex/figure_quality_sup/quality_wan_161}
\input{figure_tex/figure_quality_sup/quality_hunyuan_321}
\input{figure_tex/figure_quality_sup/quality_hunyuan_161}

%% file: figure_tex/context_length_ood.tex
\begin{figure}[h]
  \centering
  \includegraphics[width=0.95\linewidth]{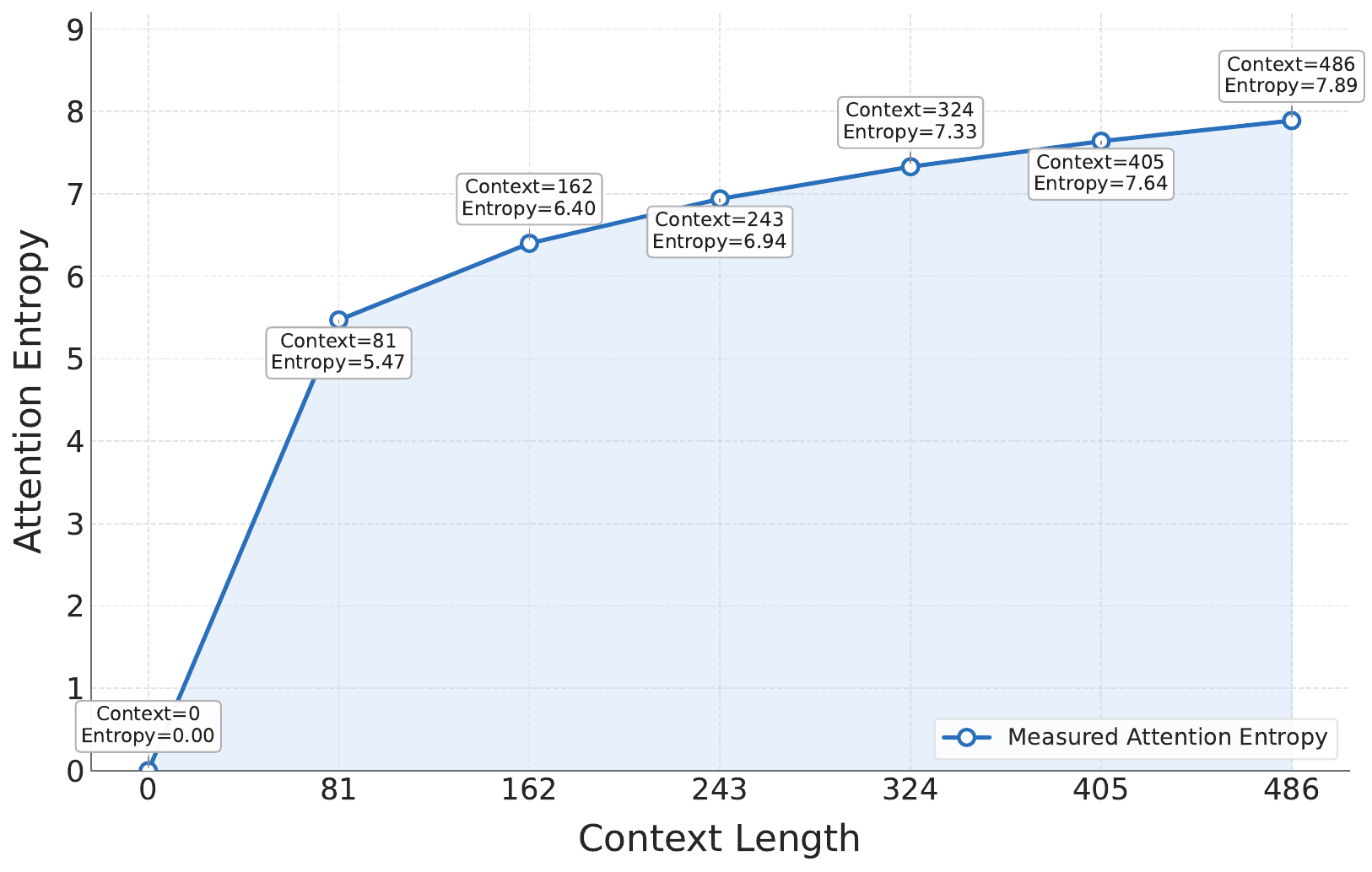}
  \vspace{-8pt}
   \caption{
   \textbf{Layer-wise sensitivity to context-length O.O.D measured via attention entropy differences.}
   }
   \label{fig:contex_length_ood}
\end{figure}

%% file: figure_tex/attention_mask.tex
\begin{figure}[t]
  \centering
  \vspace{-10pt}
  \includegraphics[width=1.05\linewidth]{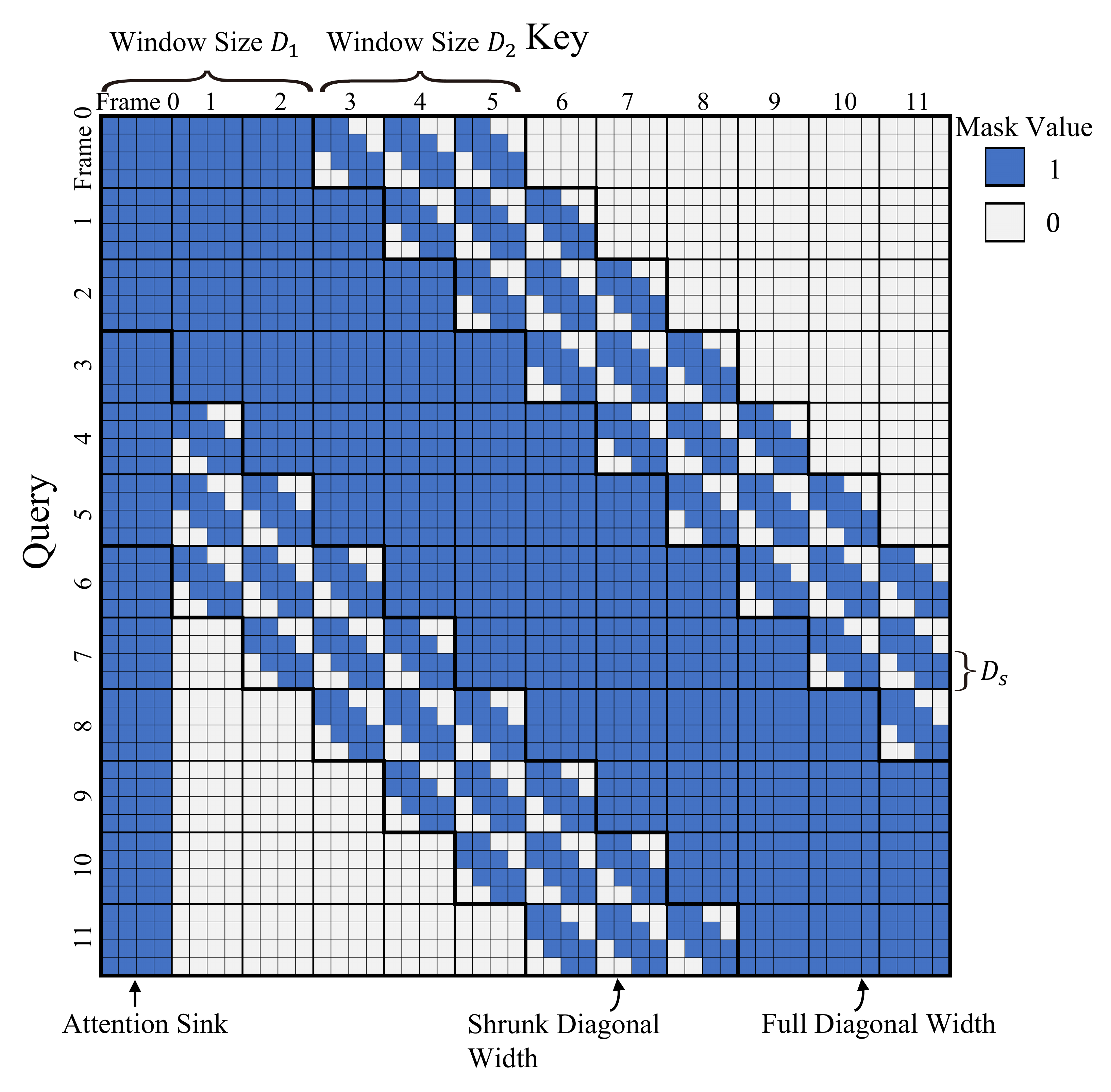}
   \caption{\textbf{Attention mask used in TSA}.}
   \label{fig:attn_mask}
   \vspace{-10pt}
\end{figure}

%% file: figure_tex/figure_quality_sup/quality_wan_321.tex
\begin{figure*}[t]
  \centering
  \vspace{-5pt}
  \includegraphics[width=1\linewidth]{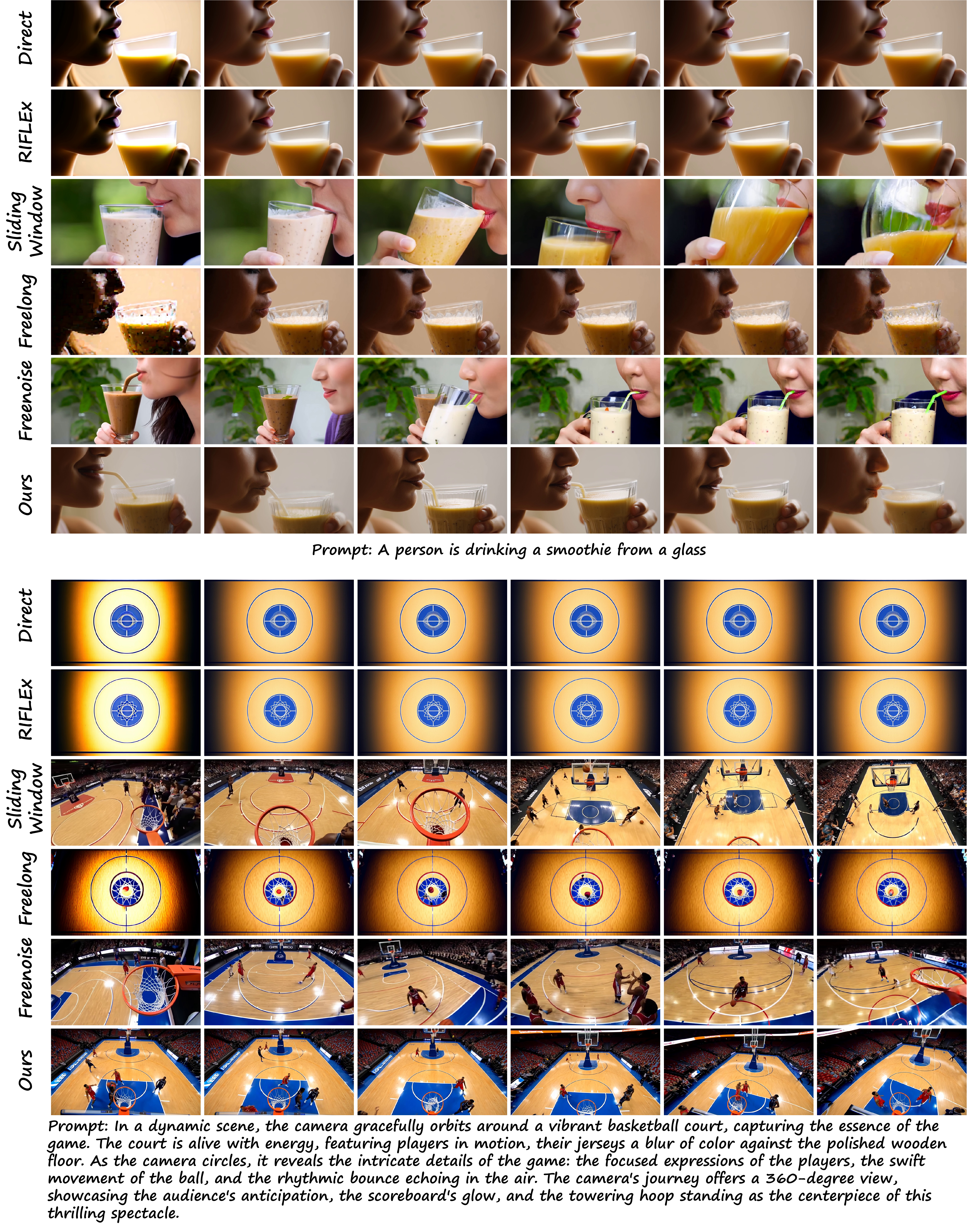}
   \caption{\textbf{Qualitative Results of Wan2.1-1.3B-T2V with $4\times$ extension (321-frame)}}
   \label{fig:quality_wan_321}
   \vspace{-10pt}
\end{figure*}

%% file: figure_tex/figure_quality_sup/quality_wan_161.tex
\begin{figure*}[t]
  \centering
  \vspace{-5pt}
  \includegraphics[width=1\linewidth]{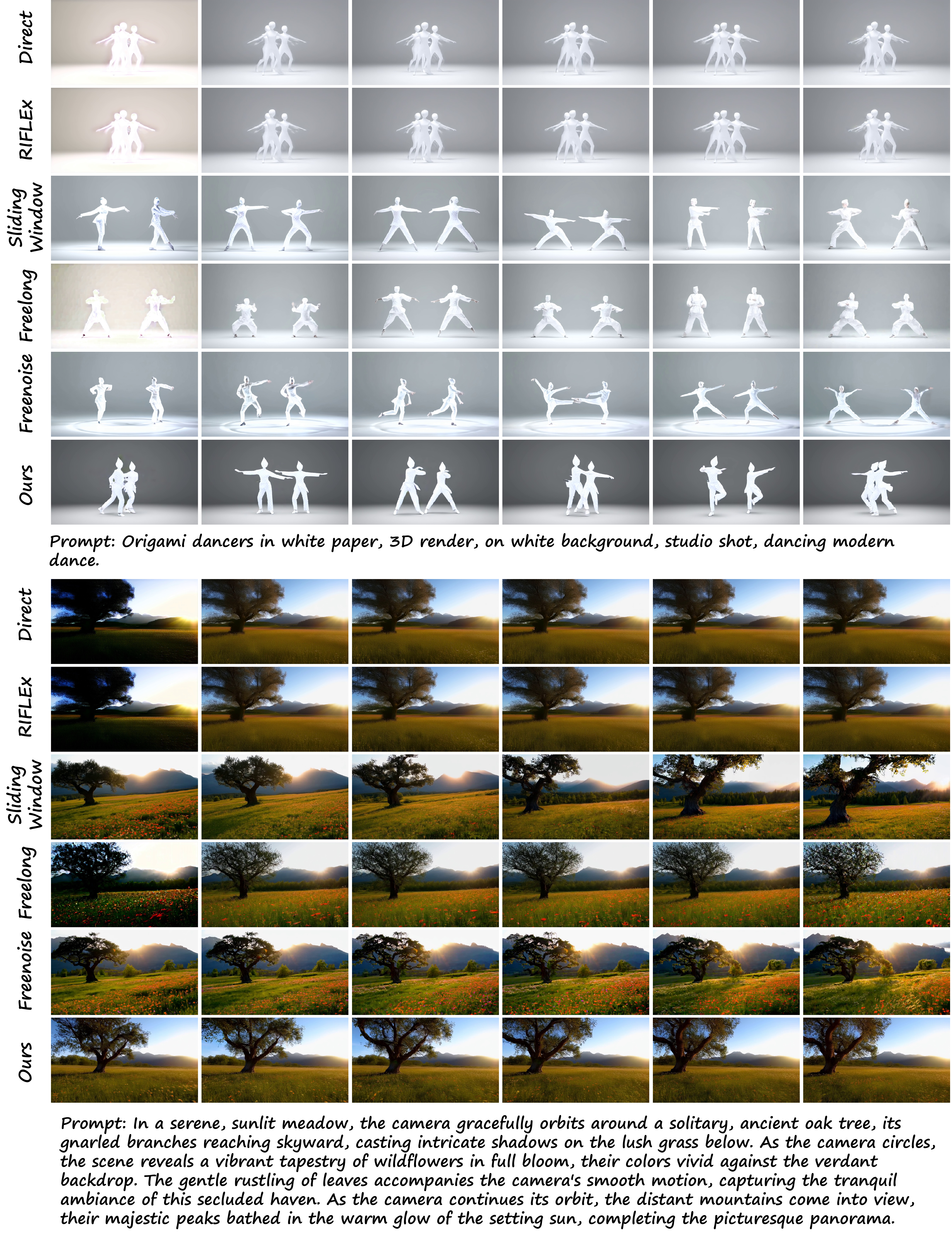}
   \caption{\textbf{Qualitative Results of Wan2.1-1.3B-T2V with $2\times$ extension (161-frame)}}
   \label{fig:quality_wan_161}
   \vspace{-10pt}
\end{figure*}

%% file: figure_tex/figure_quality_sup/quality_hunyuan_321.tex
\begin{figure*}[t]
  \centering
  \vspace{-5pt}
  \includegraphics[width=1\linewidth]{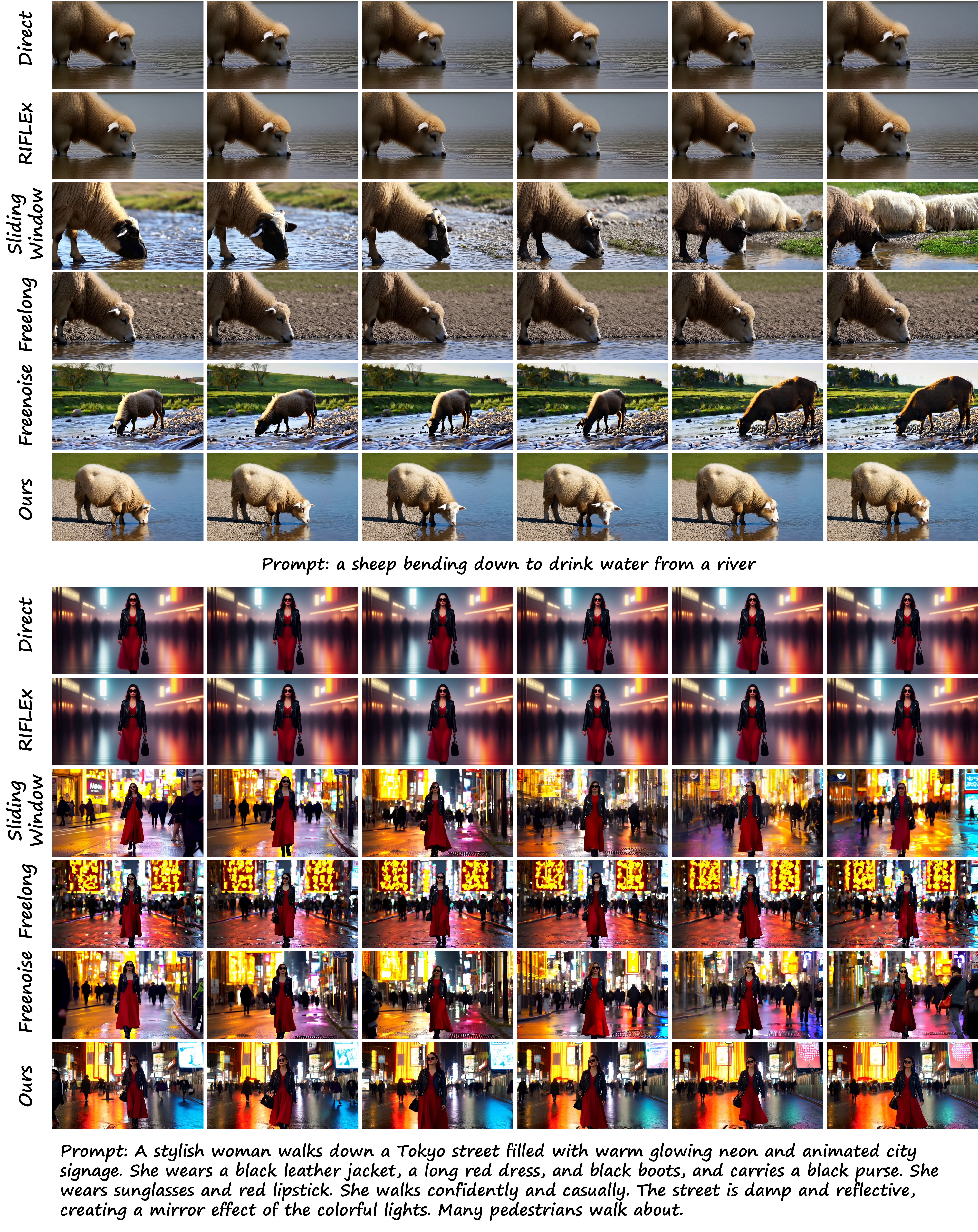}
   \caption{\textbf{Qualitative Results of HunyuanVideo with $4\times$ extension (509-frame)}}
   \label{fig:quality_hunyuan_321}
   \vspace{-10pt}
\end{figure*}

%% file: figure_tex/figure_quality_sup/quality_hunyuan_161.tex
\begin{figure*}[!th]
  \centering
  \vspace{-5pt}
  \includegraphics[width=1\linewidth]{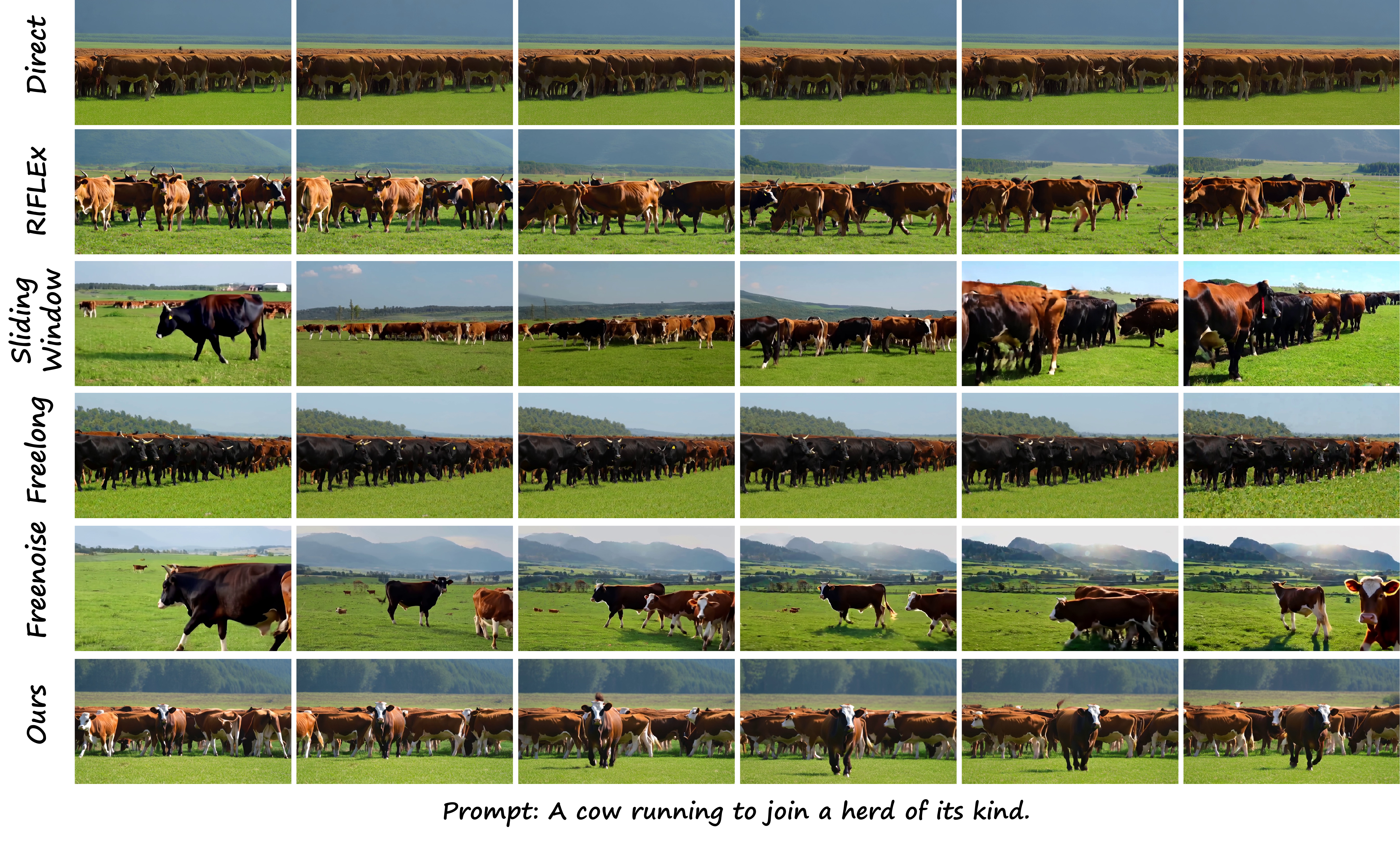}
   \caption{\textbf{Qualitative Results of HunyuanVideo with $2\times$ extension (253-frame)}}
   \label{fig:quality_hunyuan_161}
   \vspace{-10pt}
\end{figure*}